\documentclass[twoside,11pt]{article}

\usepackage[abbrvbib, preprint]{jmlr2e}

\firstpageno{1}

\usepackage{times,latexsym}
\usepackage{url}
\usepackage{amsfonts}
\usepackage[T1]{fontenc}
\usepackage{multirow}
\usepackage{tabularx}
\usepackage{booktabs}       
\usepackage{graphicx}
\usepackage{amsmath}
\usepackage{subcaption}
\usepackage{caption}
\usepackage{xcolor}         
\usepackage{algorithm}
\usepackage[noend]{algpseudocode}

\usepackage{xspace,mfirstuc,tabulary}

\definecolor{midnightgreen}{rgb}{0.0, 0.29, 0.33}
\definecolor{darkpink}{rgb}{0.91, 0.33, 0.5}

\newcommand{\model}{METRO-LM\xspace}
\newcommand{\framework}{METRO\xspace}
\newcommand{\RR}{\mathbb{R}} 

\newcommand{\bs}[1]{\boldsymbol{#1}}

\date{}

\begin{document}

\title{METRO: Efficient Denoising Pretraining of Large Scale Autoencoding Language Models with Model Generated Signals}

\author{\name Payal Bajaj\thanks{Equal contribution.} \email payal.bajaj@microsoft.com \\ \vspace{-5mm}
      \AND
      \name Chenyan Xiong$^*$ \email chenyan.xiong@microsoft.com \\ \vspace{-5mm}
      \AND
      \name Guolin Ke \email guolin.ke@microsoft.com \\ \vspace{-5mm}
      \AND
      \name Xiaodong Liu \email xiaodl@microsoft.com \\ \vspace{-5mm}
      \AND
      \name Di He \email dihe@microsoft.com \\ \vspace{-5mm}
      \AND
      \name Saurabh Tiwary \email satiwary@microsoft.com \\ \vspace{-5mm}
       \AND
      \name Tie-Yan Liu \email tyliu@microsoft.com \\ \vspace{-5mm}
       \AND
             \name Paul Bennett \email paul.n.bennett@microsoft.com \\ \vspace{-5mm}
       \AND
      \name Xia Song\email xiaso@microsoft.com \\ \vspace{-5mm}
       \AND
      \name Jianfeng Gao \email jfgao@microsoft.com \\ 
      \addr Microsoft, One Microsoft Way, Redmond, WA 98052
      }

\editor{}

\maketitle

\begin{abstract}
We present an efficient method of pretraining large-scale autoencoding language models using training signals generated by an auxiliary model. 
Originated in ELECTRA, this training strategy 
has demonstrated sample-efficiency to pretrain models at the scale of hundreds of millions of parameters.
In this work, we conduct a comprehensive empirical study, and propose a recipe, namely ``Model generated dEnoising TRaining Objective'' (METRO), which incorporates some of the best modeling techniques developed recently to speed up,  stabilize, and enhance pretrained language models without compromising model effectiveness.
The resultant models, \model{}, consisting of up to 5.4 billion parameters, achieve new state-of-the-art on the GLUE, SuperGLUE, and SQuAD benchmarks.
More importantly, \model{} are \emph{efficient} in that they often outperform previous large models with significantly smaller model sizes and lower pretraining cost.
\end{abstract}

\section{Introduction}

With the advancement of accelerating hardware, parallel training infrastructure, and optimization techniques, the size of pretrained language models has increased at an exponential scale: from hundreds of millions to trillions in the past five years~\citep{raffel2019t5, brown2020language, smith2022MTNLG}. On one hand, larger scale language models have shown stronger generalization capability in various downstream tasks~\citep{wang2018glue, raffel2019t5}. On the other hand, it is widely observed that an exponential increase in model size and computing cost is required to improve model effectiveness at a linear rate~\citep{kaplan2020scaling}. This is concerning as the investment in computing resource is likely to yield diminishing returns and the trend of model scale growth is unlikely sustainable~\citep{strubell2019energy}.

Recent studies have revealed two main challenges for efficiently scaling up Large-scale Language Models (LLMs). The first challenge is due to the use of \textit{static training} methods, where the signals for pretraining LLMs are randomly generated using a set of predefined rules (e.g., random word masking). These methods have proven to be ineffective in improving the performance of LLMs in late stages of pretraining~\citep{clark2020electra, meng2021coco}. The other challenge is \textit{optimization stability}. Deep networks are prone to instability in the stochastic gradient optimization (SGD) process, leading to frequent divergence in large-scale pretraining. Many techniques have been proposed to improve training stability, but often with trade-offs on training efficiency and the effectiveness of the resultant models~\citep{xiong2020layer, liu2020understanding, shleifer2021normformer}. 
In addition, while the scale of auto-regressive uni-directional language models has grown by orders of magnitude, e.g., GPT-3~\citep{brown2020language} and  MT-NLG~\citep{smith2022MTNLG}, the research progress of scaling bi-directional autoencoding language models, e.g., BERT-style models, is much slower, even though autoencoders are much more widely used on many natural language understanding tasks~\citep[e.g.,][]{rajpurkar2016squad, wang2018glue, wang2019superglue}.

\begin{figure}[t]
\centering

\includegraphics[width=0.60\textwidth]{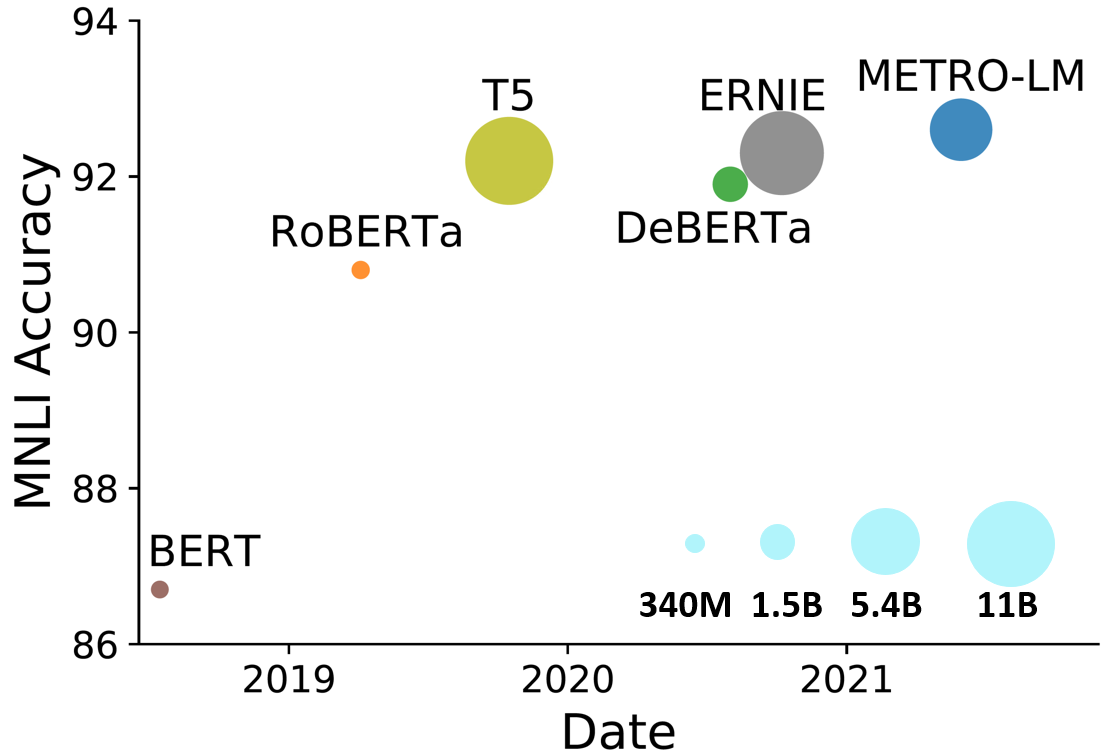}
\caption{The growth of autoencoding language models and their performances on MNLI test set. 
\vspace{-1em}
}\label{fig:intro_results}
\end{figure}

In this paper, we present an efficient denoising pretraining method for large-scale autoencoding language models. 
Unlike previous works on building larger models using more computing resources,
we focus on improving model effectiveness (parameter efficiency) and reducing computing cost (energy efficiency) of pretraining LLMs. 
As a result, we have developed a series of language models that outperform previous LLMs on a wide range of natural language understanding tasks, and, more importantly, use smaller model sizes and significantly less training cost.

Specifically, we improve modeling efficiency using two sets of techniques.
The first is to pretrain LLMs using model-generated signals. 
Instead of random masking~\citep{devlin2019bert},
we corrupt each input word sequence by replacing some of its tokens with the ones generated by an auxiliary language model. 
Then, the autoencoding language model is pretrained to \emph{denoise} the model-corrupted input.  
Originated in ELECTRA~\citep{clark2020electra}, the auxiliary model, pretrained side-by-side with the autoencoding language model, generates corrupted word sequences (training signals) with increasing difficulty to denoise as pretraining progresses, providing an effective learning curriculum for better pretraining efficiency. 
Second, we incorporate a suite of techniques of improving the efficiency and stability of large scale model training, such as the ZeRO optimizer~\citep{rajbhandari2020zero}, 
scaled initialization techniques, a set of customized Fused Operations in mix-precision training. These techniques enable us to retain model architecture designs that may cause instability in large-scale training~\citep[e.g.,][]{liu2020understanding} but beneficial to model effectiveness, such as post LayerNorm and deeper networks.

We conduct a comprehensive empirical study of language model pretraining techniques and propose a training recipe for efficient denoising pretraining of large scale autoencoding language models, namely \emph{Model-generated dEnoising TRaining Objective} (METRO). 
As shown in Table~\ref{tab:summary}, METRO significantly extends the ELETRA-style denoising training method by incorporating a suite of modeling techniques ranging from Transformer architectures, pretraining objectives, to efficient, stable, and scalable model optimization.
Using METRO, we are able to train a series of autoencoders, \model{}, the biggest instance of which consists of 5.4 billion parameters with 64 Transformer layers, in a much more parameter and energy efficient manner than previous LLMs~\citep{shoeybi2019megatron, raffel2019t5, sun2021ernie}. As shown in Figure~\ref{fig:intro_results}, \model{}  outperforms previous LLMs that consist of nearly twice model parameters. As discussed in later sections, METRO training is energy efficient in that \model{} often reaches the same performance as other SOTA LLMs using only half of their pretraining steps, while being smaller and cheaper with fewer parameters.

To facilitate the adaptation of LLMs to downstream tasks,
we also present a new model fine-tuning recipe based on posterior differential regularization~\citep{cheng2020pdr} and multi-task learning~\citep{liu2019mt-dnn}.
Our experiments show that the recipe leads to effective and stable fine-tuning for billion-parameter LLMs. The fine-tuned \model{} achieves human-level performance on the MNLI task, and creates new SOTA on both GLUE and SuperGLUE test sets. 


\begin{table}[t]
\centering
\small 
    \resizebox{\textwidth}{!}{
\begin{tabular}{ll}
\toprule
\textbf{Group} & 
{\textbf{Technique}} \\ \hline
\multirow{5}{*}{\textbf{Transformer Architecture}} &  Shallow and Wide Auxiliary Model~\citep{meng2021coco} \\
 & Post LayerNorm~\citep{devlin2019bert} \\ 
 & Relative Position Embedding~\citep{raffel2019t5} \\
 & TUPE Decoupled [CLS] Position Embedding~\citep{ke2020tupe} \\
 & Constrained Embedding Sharing~\citep{he2020deberta} \\ \hline 
\multirow{4}{*}{\textbf{Pretraining Objective}} & Model-based Corruption~\citep{clark2020electra} \\
 & Masked Language Modeling (MLM) for Auxiliary~\citep{devlin2019bert} \\
 & Replaced Token Detection (RTD) for Main~\citep{clark2020electra} \\
 & Simplified Corrective Language Modeling (CLM) for Main~\citep{meng2021coco} \\ \hline

\multirow{4}{*}{\textbf{Efficiency and Stability}} & Zero-Dropout on Auxiliary \\ 
& ZeRO Optimizer for Distributed Training~\citep{rajbhandari2020zero} \\
& Fused Operations in CUDA for Efficient Mixed Precision Training \\
& Scaled Initialization for Optimization Stability \\ \bottomrule
\end{tabular}
}
\caption{
Summary of techniques included in METRO to pretrain large scale autoencoding language models more effectively and efficiently. 
}
\label{tab:summary}
\end{table}

We structure this paper as follows.
After discussing related work in Section 2, we present in detail our construction of the METRO recipe. Section 3 presents the framework of \model{}, the denoising training method, and the experiments in the base model setup. Section 4 discusses how we scale \model{} up to billions of parameters. Section 5 and Section 6 evaluate the performance of language models of larger sizes. Section 7 concludes the paper.

\section{Related Work}

Pretrained language models (PLMs) can be grouped into two categories: \textit{autoencoding} and \textit{autoregressive} models. While both are language representation models in that they encode input text with dense vector representations, they have been used for different types of language tasks. 
Autoencoding models are widely used for language understanding tasks, such as text classification~\citep{minaee2021deep}, extractive question answering~\citep{rajpurkar2016squad}, text matching~\citep{xiong2020approximate}, information seeking~\citep{gao2022neural}. 
Autoregressive, decoder-only, models are mainly used for language generation tasks~\citep{radford2019language}. 

The effectiveness of large-scale PLMs has been demonstrated in a wide range of language tasks~\citep[e.g.,][]{devlin2019bert,liu2019roberta,yang2019xlnet,raffel2019t5, he2020deberta}.
One interesting observation is that the growth scale of dense autoregressive models is far ahead of autoencoding models. Large unidirectional models, such as GPT-3 (175B)~\citep{brown2020language}, Gopher (280B)~\citep{rae2021scaling}, and Megatron-Turing NLG (530B)~\citep{smith2022MTNLG}, are significantly bigger than the largest bi-directional autoencoding models, e.g., DeBERTa (1.5B)~\citep{he2020deberta}, Megatron (3.9B)~\citep{shoeybi2019megatron}, and ERNIE 3.0 (10B)~\citep{sun2021ernie}. 
In this work we aim to scale up autoencoding models, which are widely used for real-world language representation scenarios.

Studies on the scaling law of language models suggest that an exponential increase of investment in compute resource and model size only yields linear performance gain downstream~\citep{kaplan2020scaling}. 
Improving modeling efficiency becomes increasingly more important as LLMs get bigger and bigger.
ELECTRA~\citep{clark2020electra} presents an efficient way to pretrain an autoencoder using training signals generated by an auxiliary model. \citet{meng2021coco} pointed out that the efficiency of ELECTRA pretraining is attributed to the learning curriculum provided by the auxiliary model, as the model generated training signals are increasingly difficult to denoise during the course of pretraining. Although the ELECTRA method has been proved cost-effective in pretraining models with hundred millions of parameters, its efficiency remains unexplored at the scale of billions of parameters.

Another widely observed bottleneck in training LLMs is the optimization instability. The sources of instabilities include layer normalization placement~\citep{xiong2020layer, shleifer2021normformer}, parameter initialization~\citep{liu2020understanding}, mix precision optimization~\citep{rae2021scaling}, to name a few. 
In order to pretrain LLMs, it is common to resort to techniques that compromise model effectiveness for training stability. For example, models using a pre-LayerNorm architecture are less effective than the models using post-LayerNorm, but the former are much easier to train at large scale than the latter~\citep{liu2020understanding, xiong2020layer}. In this study we show that training stability can be achieved with fewer sacrifices on model effectiveness.

There are other pathways to improving LLM efficiency.
One is to augment a language model with a retrieval component to fetch external knowledge that is useful to perform downstream tasks. So, the size of the language model can be significantly reduced since it does not need to encode everything in model parameters~\citep{guu2020realm, Khandelwal2020Generalization, RETRO,gui2021kat, zhang2022retgen}.
With sparse model structures, Mixture of Experts models~\citep{artetxe2021efficient,fedus2021switch, zuo2021taming, zoph2022designing} adaptively activate a subset of model parameters (experts) for different inputs during model training and inference. 
The METRO method proposed in this paper is orthogonal to  retrieval-augmented models and sparsely activated models. Their combination is an interesting future work direction.

\section{Pretraining Models and Techniques}
\label{sec:model}
\begin{figure}[t]
\centering
\begin{subfigure}[t]{0.40\textwidth}
\centering
	\includegraphics[width=\textwidth]{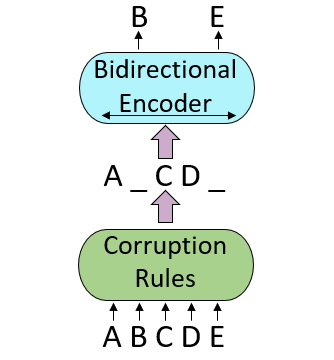}
	\caption{Rule-based Denoising\label{fig:randomdenoising}}
\end{subfigure}%
~
\begin{subfigure}[t]{0.40\textwidth}
\centering
	\includegraphics[width=\textwidth]{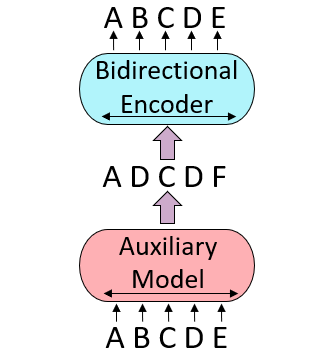}
	\caption{Model-based Denoising\label{fig:modeldenoising}}
\end{subfigure}
\caption{Training autoencoders to denoise inputs corrupted by rules versus by auxiliary models.}
\label{fig:overview}
\end{figure}

In this section we review the denoising pretraining method,  explore various techniques to construct our recipe of pretraining an autoencoding language model with one hundred million parameters, \model{}$_\text{base}$, and present experimental results at the \textit{base} setting.

\subsection{Denoising Pretraining}

The general denoising pretraining setup can be written as
\begin{align}
    X^\text{orig} \xrightarrow{\text{corruption}} X^\text{noise} \xrightarrow{\text{model}} X^\text{orig}.
\end{align}
The text input $X^\text{orig}$ is first corrupted to form the (noisy input) $X^\text{noise}$, upon which the model is trained to recover the original sequence.

Denoising training is a self-supervised learning method in that the training signals can be generated automatically in arbitrarily large quantity. Its flexibility also makes it handy to pretrain bi-directional autoencoders which often uses non-autoregressive training~\citep{devlin2019bert}. 
Denoising training methods can be grouped into two categories depending on whether the input sequence is corrupted based on rules or machine learned models.

\paragraph{Rule-based Corruption.} A simple way to construct $X^\text{noise}$ is to use a set of rules to randomly corrupt some of its tokens:
\begin{align}
  X^\text{orig} \xrightarrow{\text{random corruption by rules}} X^\text{noise}.   
\end{align}
As illustrated in Figure~\ref{fig:randomdenoising}, the Mask Language Modeling (MLM) task~\citep{devlin2019bert} randomly replaces a fraction of tokens (e.g., 15\%) in $X^\text{orig}$ with the special token ``[MASK]'': $X^\text{noise} = [x_1^\text{orig},...,\text{[MASK]}_i,...x_n^\text{orig}]$. 
MLM proves to be one of the most effective rules for pretraining denoising LLMs~\citep{raffel2019t5}.

\paragraph{Model-based Corruption.} 
One limitation of rule-based denoising pretraining is that the randomly generated noises are not informative enough for effective modeling in the later stages of pretraining, when models already learned certain language semantics~\citep{clark2020electra, meng2021coco, meng2022amos}. ELECTRA~\citep{clark2020electra}, using model-based corruption, is among the most effective methods to pretrain denoising autoencoders. 
As illustrated in Figure~\ref{fig:modeldenoising}, it employs an auxiliary Transformer to construct $X^\text{noise}$, upon which the main model is trained to (partially) recover the original sequence:
\begin{align}
      X^\text{orig} &\xrightarrow{\text{auxiliary model}} X^\text{noise}\xrightarrow{\text{main model}} \Tilde{X}^\text{orig}.
\end{align}
The auxiliary model is often a language model, such as a BERT-style model pretrained using MLM. The corruption is conducted by replacing a fraction of original tokens with samples $x_i^\text{mlm}$, generated by the auxiliary language model as $X^\text{noise}=\{x_1^\text{orig},...,x_i^\text{mlm},...x_n^\text{orig}\}$.
The denoising objective that guides the training of the main model ranges from a binary classification objective of whether the token in $X^\text{noise}$ is the original~\citep{clark2020electra}, or a multi-choice question of selecting the original token from a set of model-generated samples~\citep{xu2020mc}, to recovering $X^\text{orig}$~\citep{meng2021coco}. 
In this study we follow the terminology used in \citet{meng2021coco} and refer to the two models as \emph{auxiliary} and \emph{main} models, respectively, instead of \emph{generator} and \emph{discriminator} as in \citet{clark2020electra}, because the main model can also be a generative language model.

The major advantage of model-based denoising training is the learning curriculum formed by the side-by-side pretraining of the two models. The auxiliary model, starting from random weights, learns to produce noisy samples $X^\text{noise}$ that are increasingly challenging to denoise by the main model.
The benefit of such a learning curriculum has been demonstrated in many recent studies, but 
only for models whose parameters are in the hundreds of millions, i.e., in BERT base and large settings~\citep{xu2020mc, shen2021training, meng2021coco, meng2022amos}. 
In this work, we extend previous studies and propose the Model-generated dEnoising TRaining Objective (METRO), a denoising training recipe for pretraining LLMs with billions of parameters.

\subsection{Construction of \model{}}

To scale up denoising training using \model{}, we need to study different model configurations among the recently proposed ELECTRA-style models and various language model pretraining techniques. Due to the high computing cost, conducting a thorough empirical study on models with billions parameters is prohibitively expensive~\citep{raffel2019t5, brown2020language}. 
A more feasible approach is to explore using the BERT base setup, identify the best combination of these techniques that achieve a good tradeoff among efficiency, effectiveness, simplicity, and stability, and then scale it up to larger sizes.

In the rest of this section we describe in detail the baseline METRO model, which is used as a backbone to study various modeling configurations and techniques, and our eventual model \model{}$_\text{Base}$.
In the next section we describe how we scale up to billions parameters.

\subsubsection{Baseline Model}

Our \textit{baseline} model follows the vanilla training setup of ELECTRA~\citep{clark2020electra}, using a MLM language model as the auxiliary, and the Replaced Token Detection (RTD) task to train the main model.

\paragraph{Baseline Training Objectives.} 
Given an input sequence with 15\% tokens masked,
 $X^\text{mask}=[x_1^\text{orig},...,\text{[MASK]}_i,...,x_n^\text{orig}]$,
the \textit{auxiliary} model $g_\text{aux}$ is trained using the standard MLM task:
\begin{align}
       \mathcal{L}_\text{aux} = \mathbb{E} ( - \sum_{i\in \mathcal{M}} \log  p_{\text{mlm}} ( x_i^\text{orig} \big| \bs{h}^\text{aux}_i )); \text{ } p_\text{mlm}(x_t|\bs{h}^\text{aux}_i) = \frac{\exp(\bs{x}_t^T \bs{h}^\text{aux}_i)}{\sum_{t'=1}^{|V|}\exp(\bs{x}_{t'}^T \bs{h}^\text{aux}_i)}.
\end{align}
The auxiliary Transformer $g_\text{aux}$ first produces the contextualized representation $\bs{h}_i^\text{aux}$ for each token, which is used by the language model head $p_\text{mlm}$ to generate a softmax distribution over the vocabulary. The auxiliary model is pretrained to recover the masked tokens in $\mathcal{M}$ using the standard language modeling loss $\mathcal{L}_\text{aux}$.

Then the auxiliary model is used to corrupt text sequences for the denoising pretraining objective for the main model training. Following ELECTRA, we construct $X^\text{noise}$ by sampling tokens from $g_\text{aux}$ at all masked positions:
\begin{align}
 x_i^\text{noise} &\sim p_{\text{mlm}} \left(x | \bs{h}^\text{aux}_i \right),\, \text{if $i \in \mathcal{M}$ }; \text{ }  x_i^\text{noise} = x_i^\text{orig},\, \text{else.} \label{eq:corrupt}
\end{align}

The main model, Transformer $g_\text{main}$, is then trained using the binary RTD task: 
\begin{align}
    \mathcal{L}^\text{rtd}_{\text{main}} = \mathbb{E} ( -&\sum_{x_i^\text{noise} = x_i^\text{orig}} \log p_{\text{rtd}}(x_i^\text{noise} = x_i^\text{orig} \big| \bs{h}_i) \nonumber \\
     -&\sum_{x_i^\text{noise} \neq x_i^\text{orig}} \log (1 - p_{\text{rtd}}(x_i^\text{noise} = x_i^\text{orig} \big| \bs{h}_i ))). \label{eq:rtd}
\end{align}
It learns to detect whether a token $x_i^\text{noise}$ is being replaced (noise) or not (original), using  $p_\text{rtd}$, a binary classification head upon $\bs{h}_i$, the contextualized representation generated by $g_\text{main}$.

The auxiliary model and the main model are pretrained side-by-side as
\begin{align}
\mathcal{L} = \mathcal{L}_{\text{aux}} + \lambda \mathcal{L}^\text{rtd}_{\text{main}}.  \label{eq:total_loss}
\end{align}
The auxiliary model learns to construct samples that are increasingly harder to  denoise, thus forms an effective learning curriculum for the main model. The hyperparameter $\lambda$ balances the learning speeds of the main and auxiliary models.

\paragraph{Baseline Configurations.} 
We improve the vanilla ELECTRA in our baseline model with several modeling practices whose effectiveness is verified in recent studies. 

\begin{itemize}
    \item \textit{Shallow auxiliary model.} 
    Following COCO-LM~\citep{meng2021coco}, we keep the same hidden dimension with the main model but reduce the network depths of the auxiliary model, e.g., to 1/3 in base settings. 
    \item \textit{Relative Position Embedding.} 
    Following T5~\citep{raffel2019t5}, we use relative position bins for both the auxiliary and main model, with 32 bins and 128 maximum distance.
    \item \textit{Large Vocabulary.} ~\citet{unilmv2} show that larger vocabulary size improves the capability of LLMs without much cost in training or inference speeds. Following DeBERTa~\citep{he2020deberta}, we use a cased sentence piece BPE vocabulary of 128K tokens. 
    \item \textit{Respecting Document Boundary.} We respect document boundary and do not concatenate texts from different documents into one pretraining sequence. 
\end{itemize}

\subsubsection{Candidate Techniques to Integrate}

Recent studies have developed various modeling techniques in pretraining. The benefits of these techniques are not as well-verified as, for example, relative position embedding. Many of them also provide different trade-offs between effectiveness and efficiency.  To identify the best strategy of scaling up \model{},
we explore in the \framework{} pretraining framework three categories of modeling techniques: 
Transformer architectures, training objectives and dropout usage. 
In what follows, we brief these techniques and refer readers to the original papers for details.

\paragraph{Transformer Architectures.} 
There has been debate on whether the power of large scale pretraining overwhelms the inductive biases introduced via network architecture design. Empirically, the vanilla Transformer with relative position embedding often outperforms more sophisticated variations~\citep{narang2021transformer}. 
Nevertheless, certain architecture designs provide good trade-off between capacity and stability. In this work, we explore three groups of techniques on architecture changes: LayerNorm location, position embeddings, and embedding sharing between auxiliary and main models.

\begin{itemize}
    \item \textit{LayerNorm Location.} Besides the post-Layernorm used in our baseline, we also experiment with pre-Layernorm~\citep{xiong2020layer}, where Layernorm is performed before self-attention and FFN.
    \item \textit{Position Embeddings.} 
    We experiment to increase the relative position embedding bins from 32 to 64. We also explore the variant position embedding configurations developed in TUPE~\citep{ke2020tupe}, including decoupling the position embedding correlations from word embeddings and resetting the position embedding of [CLS].
    \item \textit{Constrained Embedding Sharing.} In ELECTRA-style models, by default the word and position embeddings used for the auxiliary and the main are shared. Inspired by 
    DeBERTa~\citep{he2020deberta}, we experiment two different levels of constrained embedding sharing between the two models: using \textit{individual position embeddings} and using \textit{individual LM bias terms} in language modeling heads. The latter is used only when the main model is trained via language modeling objectives. This grants the model slightly more flexibility.
\end{itemize}

\paragraph{Training Objectives.} 
Several new training objectives have been proposed to improve the RTD task for the main Transformer, aiming to recover more information of the original sequence than detecting whether a token is replaced.

\begin{itemize}
    \item \textit{Replace MLM} trains the main model to predict the original tokens at masked positions. It differs from the standard MLM in that the input tokens are sampled from the auxiliary model. It is first experimented in ELECTRA~\citep{clark2020electra} in place of RTD. 
    
    \item \textit{One-choice cloze} trains the main model to pick the original token from three randomly generated tokens at masked positions~\citep{xu2020mc}. This task is more challenging than RTD but easier than Replace MLM which requires prediction from the entire vocabulary.
    
    \item \textit{Corrective Language Modeling (CLM)} provides a multi-task setup that combines RTD with the language modeling task~\citep{meng2021coco}. It trains the main model to both classify whether each position is replaced using the RTD head and predict the original token using a language modeling head. The language modeling head also includes a copy mechanism, which reuses the RTD head. A stop-gradient operation is introduced to prevent the simpler RTD task being distracted by the harder language modeling task.
    
    \item \textit{Simplified CLM.} We propose a simplified version of CLM 
    by only applying the CLM task on the masked positions $\mathcal{M}$ and omitting the copy mechanism, as 
    \begin{align}
        \mathcal{L}^\text{s-clm}_\text{main} = \mathbb{E} ( - \sum_{i\in \mathcal{M}} \log  p_{\text{clm}} ( x_i^\text{orig} \big| \bs{h}_i )); \text{ } p_\text{clm}(x_i | \bs{h}_i) = \frac{\exp(\bs{x}_i^\top \bs{h}_i)}{\sum_{x_t \in V}\exp(\bs{x}_t^\top \bs{h}_i)}.
    \end{align}
    It is trained in a multi-task setup with the RTD task by linearly combining two losses:
    \begin{align}
        \mathcal{L}_\text{main} &= \lambda \mathcal{L}^\text{rtd}_\text{main} + \mathcal{L}^\text{s-clm}_\text{main}. \label{eq:s-clm}  
    \end{align}
    \item \textit{Sequence Contrastive Learning (SCL)} is a sequence-level task that trains the Transformer to align related sequence pairs closer than unrelated ones in the embedding space. 
    When it was first introduced in COCO-LM~\citep{meng2021coco}, the related sequence pairs are constructed by simple data augmentation (cropping) and the unrelated ones are sampled within the mini-batch. Other forms of related sequences have been explored in the continuous pretraining and pre-finetuning setup~\citep{ni2021large, neelakantan2022text}. In this work, we experiment SCL following the COCO-LM setup.
\end{itemize}

\paragraph{Dropout.}
We also experiment the \emph{zero-dropout} method on the auxiliary side where we disable the dropout regularization completely. It significantly reduces the computing cost as both the training of the auxiliary model and its inference (to construct training sequences for the main) can be done in one forward-backward pass, in contrast to two separate passes as in \citet{meng2021coco}.

\subsubsection{Empirical Explorations for Best \model{} Configuration}

Given the baseline model and the set of candidate modeling techniques described above, we now explore how to best combine them for a configuration of \model{} that can be most effectively and efficiently scaled up to billions of parameters. 

The technical choices for billion-parameter models are dramatically different and more constrained. For example, a technique that increases downstream performance by 3\% at the cost of 30\% more parameters is acceptable for base models, but is considered prohibitive expensive at larger sizes.
The increased optimization instability at larger scale also makes it challenging to use techniques that increase model effectiveness at the cost of reducing stability. Pretraining on thousands of GPUs for several weeks but yielding a deficient model due to divergence is highly undesirable. 
Therefore, we explore the trade-off among effectiveness, computing cost, and stability/simplicity by mainly running experiments in the base model setting, as an approximation to potential model behaviors at large scale.

\paragraph{Experiment Settings.} 
We conduct explorations in the standard BERT base setup, with the goal of identifying the configuration that yields the best trade-off among effectiveness, efficiency, and stability. Specifically, we follow the \textit{base} experimental setting used in previous studies~\citep{devlin2019bert, clark2020electra,meng2021coco}. 
We pretrain BERT$_\text{base}$ style 12-layer main Transformer with 768 hidden dimension on Wikipedia + Book Corpus, with sequence length 512, batch size 2048, and for 125K steps. The auxiliary model shares the same configuration with the main model, but uses 4 Transformer layers. 
Following recent practices~\citep{yang2019xlnet, liu2019roberta, lan2019albert},
we use single-task, single-model fine-tuning results on MNLI and SQuAD 2.0 to measure model effectiveness and closely track model efficiency and stability. We also measure models' average score on GLUE\footnote{Note that the results of tasks like CoLA and MRPC have high variance and need to be taken with grain of salt.}.


\begin{table}[t]
\centering

\small 
\begin{tabular}{lccc}
\toprule
& \multicolumn{2}{c}{\textbf{GLUE}} & \textbf{SQuAD 2.0} \\ \cline{2-4}
& \textbf{MNLI-m/mm} & \textbf{AVG Large/All} & \textbf{EM/F1} \\
\midrule
ELECTRA
&  86.9/86.7 & 91.2/85.5 & 80.5/83.3\\
COCO-LM & 88.5/88.3 & 91.7/87.2 & 82.4/85.2 \\
Baseline
& \textbf{89.0}/88.5	& 92.1/88.3 &   83.8/86.7 \\
\midrule
\multicolumn{4}{l}{\textbf{Training Objectives}} \\ 
\midrule
Replace MLM  & 88.7/88.6	&92.1/88.1	&	83.6/86.5\\
One-choice cloze & 88.8/88.5	&	92.1/87.9	&	83.3/86.2\\
CLM  &  88.8/88.6 &	92.1/88.5	&	83.6/86.5	\\
RTD+SCL & 88.9/88.4 &	92.1/88.8 &	84.0/86.7 \\

\midrule
\multicolumn{4}{l}{\textbf{Transformer Architectures}} \\ 
\midrule
Pre-LayerNorm & 88.3/87.7 & 91.5/87.5 & 81.9/84.7 \\
64 Rel-Pos Bins & \textbf{89.0}/88.6 &	92.1/88.4 &	83.7/86.4 \\
TUPE Embedding & \textbf{89.0}/88.5	& 92.1/88.4  & 83.0/85.8  \\

\midrule
\multicolumn{4}{l}{\textbf{Combo: Adding techniques sequentially per row to baseline.}} \\ 
\midrule
+CLM+Zero-Dropout 
& 88.7/88.3	& 92.2/88.6  & 83.9/86.6  \\
+More Rel-Pos Bins  & 88.9/88.5	&	92.3/88.5	&	83.7/86.6\\
+ Reset [CLS]  & 88.8/88.5	&	92.2/88.4	&  84.2/87.1\\
+ Simplified CLM  & 88.8/88.6	& 92.2/88.4  & \textbf{84.4}/87.1  \\
 + Constrained Embedding Sharing & \textbf{89.0/88.8}	& \textbf{92.4/88.7}  & 84.3/\textbf{87.2}  \\
\bottomrule
\end{tabular}
\caption{
\textit{Base} model results on MNLI and SQuAD 2.0. All results are single-task and single-model evaluations on the dev set. We also report the average of four large GLUE tasks (AVG Large), MNLI, QQP, QNLI, and SST-2, as well as the average (AVG All) of large tasks with four smaller tasks: CoLA, RTE, MRPC, and STS-B. All our models are pretrained under the same setting. ELECTRA and COCO-LM numbers are from \citet{meng2021coco}. 
}
\vspace{-0.5em}
\label{tab:small_base_ablation_res}
\end{table}

\paragraph{Evaluation Results.} 
The experimental results of base models are presented in Table~\ref{tab:small_base_ablation_res}.
Our baseline model already outperforms previous state-of-the-arts by significant margins. We attribute the superior performance to the selection of modeling techniques and configuration details.

Although the four \textit{training objectives} do not provide much improvement in our experiments, CLM is the most appealing since it is more stable than Replace MLM, and the language modeling capability is required for certain downstream usages such as prompt-based learning~\citep{meng2021coco}. The benefit of SCL task is visible in sequence and document representation tasks, e.g., dense retrieval~\citep{gao2021unsupervised}, but not as much on GLUE and SQuAD.
How to better tailor pretraining to downstream tasks like dense retrieval and sentence representation is a future research direction. In this work, we omit the SCL task in scaling up as it introduces a noticeable computing cost in the COCO-LM setup.

The most visible impact on model performance due to Transformer architecture is that the accuracy drops when we switch to Pre-LayerNorm, confirming the results reported in previous studies~\citep{liu2020understanding, shleifer2021normformer}.
Yet this accuracy drop sometimes is viewed as a necessary price to pay for scaling as Pre-LayerNorm significantly improves the optimization stability, making it feasible to pretrain LLMs at larger scales. However, in this work we aim to improve optimization stability without sacrificing model effectiveness. We do not use Pre-LayerNorm in our final model. 
The various relative position embedding setups also yield ambivalent results. 

\paragraph{The Final \model{} Model.} 
To construct the final \model{}$_\text{base}$, 
we start from our very strong baseline model, add one group of techniques at a time, and integrate them into \model{}$_\text{base}$ after they are empirically verified. 
Our final search path is listed in the "Combo" group at the end of Table~\ref{tab:small_base_ablation_res}. 

Upon the baseline, our final \model{}$_\text{base}$ model uses (1) zero-dropout on the auxiliary side, which reduces computing cost, (2) TUPE's reset [CLS] embedding, which works well on SQuAD, and (3) simplified CLM, which maintains the language model capability for the main Transformer, while being more stable and simpler than COCO-LM's CLM implementation.
We also use larger relative position embedding bins, individual absolute position embeddings parameters, and separated LM head biases on the auxiliary and main models (constrained embedding sharing). These techniques improve the model's capacity without much efficiency cost.

\section{Scaling Up to Billions of Parameters}

After finding an effective configuration of \model{}$_\text{base}$, the next step is to scale it up, where a new set of challenges emerges. 
Training large models is extremely slow and computationally heavy which restricts the number of experiments that can be run with limited resources.
The size of LLMs also quickly reaches to a point where native training requires memory beyond the limit of hardware, etc. 
As models get deeper, optimization stability becomes a bigger issue due to gradient exploding and vanishing, mixed precision overflow, etc. 
An emerging unique challenge due to the use of model generated denoising objectives is the need of properly calibrating the auxiliary and the main model to construct an effective learning curriculum.

In this section, we first discuss the configurations of \model{} at larger sizes, followed by the techniques we used to improve the efficiency of distributed pretraining.
After that, we describe techniques that improve optimization stability of \model{} and summarize our findings during our scaling up attempts.
We also present our procedures to stably finetune LLMs, which is crucial for their applications on those downstream tasks without many training labels.

\subsection{Model Configuration}

\begin{table}[t]
\centering

\small 
\begin{tabular}{lll*{3}{l}ll}
\toprule
{\textbf{Model}} & {\textbf{\#Params}} & {\textbf{\#Params}} & {\textbf{Hidden}} & {\textbf{FFN}} & {\textbf{Depth}} & {\textbf{Depth}} & {\textbf{Attention}}\\ 
{\textbf{Size}} & {\textbf{(Main)}} & {\textbf{(Aux)}} & {\textbf{Size}} & {\textbf{Width}}  & {\textbf{(Main)}} & {\textbf{(Aux)}} & {\textbf{Heads}}\\ 
\midrule
Base  &  184M & 29M & 768 & 3072 &	12 &   4       &	12  \\
\midrule
Large  & 434M & 116M & 1024 & 4096   &	24  &   6    &	16  \\
\midrule
XL  & 1.6B & 300M & 1536 & 6144  &	48  &   8      &	24  \\ 
\midrule
XXL  & 5.4B & 600M & 2560 & 10240  &	64  &   8     &	 40 \\
\midrule
\end{tabular}
\caption{
Configurations of \model{}. The auxiliary model and main model only differ in their layer depths. Only the main model is used for fine-tuning. Its parameter count reflects the final model size, i.e., in terms of fine-tuning cost.
}
\label{tab:model_configs}
\end{table}

How to configure a proper learning curriculum is a key research problem in curriculum learning~\citep{graves2017automated}.  METRO is no exception.
A weak auxiliary model does not generate challenging enough training signals to optimize the main model, while a strong auxiliary model could make the denoising task a bit too hard for the main model to learn.
As the first attempt to scale up ELECTRA-style models, we keep the width of auxiliary models the same as the main,  scale up the main model by running ablations on the depth of auxiliary for a limited amount of pretraining steps, and pick the one that leads to stable learning behaviors of the main Transformer.

The configurations of \model{} at variant sizes are summarized in Table~\ref{tab:model_configs}. 
We follow the traditions at base and large from recent research~\citep{liu2019roberta, meng2021coco}. 
For XL and XXL, different from previous practices~\citep{raffel2019t5, brown2020language}, we scale the models in both width and depth. 
Though it has been observed that a deeper network often yields better generalization in deep learning, increasing the depth of LLM yields more optimization challenges, especially with Post-LayerNorm~\citep{xiong2020layer, shleifer2021normformer}. 
In this work, we aim to address the optimization stability issues, using techniques described later in this section, and to retain the effectiveness of Post Layernorm and deeper networks.

\begin{table}[t]
\centering
\small 
\begin{tabular}{lll*{2}{l}}
\toprule
{\textbf{ZeRO Stage}} & {\textbf{Params}} & {\textbf{Gradients}} & {\textbf{Optimizer}} & {\textbf{Total}} \\ 
\midrule
 None (no ZeRO) & 12G &  12G & 96G & 120G \\
\midrule
 One (Optimizer) & 12G &  12G & 400MB & 24.4G \\
\midrule
Two (Gradient) & 12G &  50MB & 400MB & 12.5G \\ 
\midrule
Three (Model) & 50MB &  50MB & 400MB & 500MB \\
\midrule
\end{tabular}
\caption{
Memory usage per GPU by each component in the optimization process of the XXL model on 256 GPUs with different ZeRO stages. Each stage adds one component (in bracket) to shard across GPUs. 
}
\vspace{-0.5em}
\label{tab:deepspeed_zero}
\end{table}

\subsection{Techniques for Training Efficiency}

We use two sets of techniques to improve distributed training efficiency: ZeRO Optimizer and Customized Fused CUDA Operations. Both impose minimum change to the model architecture and implementation.

\paragraph{ZeRO Optimizer.} In many stochastic optimizers, the size of gradients and optimizer states grows linearly with model sizes. Take Adam as an example, the optimizer state consumes more GPU memory than the model itself, making it harder to fit larger models into any single-GPU memory. 
The DeepSpeed ZeRO optimizer~\citep{rajbhandari2020zero} provides effective ways to partition the optimizer states, gradients, and model weights across multiple GPUs in distributed training, which significantly reduces the GPU memory consumption. 
Table~\ref{tab:deepspeed_zero} shows the memory requirements in different stages of ZeRO. 
Using ZeRO stage one, we are able to fit a model with in total 6B parameters 
in one A100 GPU with 40GB memory. This is because the stage one ZeRO partitions optimizer states across GPUs. E.g., for Adam, its 32-bit weights, and the first and second moment estimates are partitioned across multiple GPUs, which significantly reduces memory usage. 
This is appealing especially given the fact that this optimizer state sharing imposes no extra communication cost by redesigning the communication loop between GPUs.
We choose ZeRO stage one in our scaling up experiments due to its significant efficiency benefits and zero communication cost.

\begin{algorithm}[t]
\caption{Perform Operations in FP32 for Stability, without Fused Ops}
\begin{algorithmic}[1]
\Procedure{stable\_op\_in\_fp32}{$\bs{x}, F$}  \Comment{Input: $\bs{x}$ an tensor of size n, F the operation}
    \For {$i\ =\ 1\ to\ n$}     \Comment{Cast $\bs{x}$ to FP32}
    
        $\bs{t1}[i] = \bs{x}[i].float()$  \Comment{Creating $\bs{t1}$, an FP32 tensor of size O(n) size}
    \EndFor
    \For {$i\ =\ 1\ to\ n$}     \Comment{Perform op F with FP32}
        
        $\bs{t2}[i] = F(\bs{t1}[i])$      \Comment{Creating $\bs{t2}$, an FP32 tensor of size O(n)}
    \EndFor    
     \For {$i\ =\ 1\ to\ n$}    \Comment{Cast result to FP16}
     
        $\bs{y}[i] = \bs{t2}[i].half()$   \Comment {Creating $\bs{y}$, the output FP16 tensor of size O(n)}
    \EndFor \label{no_fused_ops}
\EndProcedure

\end{algorithmic}
\end{algorithm}

\begin{algorithm}[t]
\caption{Perform Operations in FP32 for Stability, with Fused Ops}
\begin{algorithmic}[1]
\Procedure{stable\_op\_in\_fp32}{$x, F$}   \Comment{Input: $\bs{x}$ an tensor of size n, F the operation}
    \For {$i\ =\ 1\ to\ n$}     \Comment{Performing Cast and F Element-wise}
    
        $t1 = \bs{x}[i].float()$     \Comment{t1 is a single fp32 variable}
        
        $t2 = F(t1)$            \Comment{t2 is a single fp32 variable}
        
        $\bs{y}[i] = t2.half()$      \Comment {Creating  $\bs{y}$, the output fp16 tensor of size O(n)}
        
    \EndFor  \label{with_fused_ops}
\EndProcedure
\end{algorithmic}
\end{algorithm}
\begin{table}[t]
\centering
\small 
\begin{tabular}{lll}
\toprule
{\textbf{Fused Ops}} & \textbf{Training Speed Up} & {\textbf{Memory Savings}} \\ 
\midrule
Fused Fast LN & 7\% & 0\%\\
\midrule
Fused Softmax + Dropout  & 16\% & 17\%\\ 
\midrule
\end{tabular}
\caption{
Efficiency improvements with our customized Fused CUDA operations. 
}
\vspace{-0.5em}
\label{tab:fused_ops}
\end{table}
\paragraph{Fused CUDA Operations.} 
Mixed-prediction training, where model training is performed in FP16 precision, except for the main copy of model which uses FP32 precision, is the standard for large scale training to reduce memory and communication cost.
However, it is often necessary to use FP32 for some operations to ensure numerical stability, such as \textit{softmax} in the attention mechanism. 
In existing mixed-prediction implementations, e.g., with PyTorch's \textit{amp} package, the procedure is to first cast input tensors using FP32, perform the operation, and then cast results back to FP16. This cast operation induces significant additional memory cost, as a byproduct of the abstraction needed in highlevel deep learning frameworks.

We eliminate this cost by creating customized CUDA kernels that fuse the FP32 casting and operations together, conduct the casting and computing element-wisely, and leverage computing locality for better efficiency. Algorithms 1 and 2 illustrate the difference between the mix-precision implementation with the default \textit{amp} implementation and that with our Fused CUDA operations. The two are mathematically equivalent but ours are much more efficient.
In our experiments, we use this customized Fused Ops in two places for every transformer layer: \textit{softmax} followed by the \textit{dropout} operation, which converts attention weights to attention probabilities, and all \textit{layer norms}. The memory savings and training speed-up from each operation is reported in Table~\ref{tab:fused_ops}.
In total, we achieve 23\% operation speed up and 17\% memory savings, solely from engineering optimization.

Note this engineering improvement has also been observed in~\citet{izsak2021train}.
This reflects the potential benefits of customized solutions for deep learning workloads where computing efficiency is more of a concern. A slight trade-off on implementation generality is reasonable. We have open-sourced our customized CUDA kernels to help reduce the training cost of large scale Transformers, especially for research environments with limited budgets.\footnote{https://github.com/microsoft/COCO-LM/tree/main/fairseq/fused\_ops}

\subsection{Techniques for Optimization Stability}

One of the biggest challenges in pretraining large language models is the optimization stability~\citep{liu2020understanding}. The pretraining process is prone to divergence with various gradient and numerical issues.
Our desire to scale \model{} up in depth and use Post-LayerNorm for its effectiveness makes the optimization stability even more critical.
In this section we describe the techniques of improving the optimization stability when scaling up.

There are two common ways to improve optimization stability. One is to reduce the peak learning rate, making the gradient updates more subtle. The other is to clip the gradient norm to reduce the turbulence in the stochastic optimization. 
In addition, we also scale the initialization of FFNs weights in the Transformer layers dynamically based on their depth. This helps avoid large gradients in deeper layers, thus improves the optimization stability with Post-LayerNorm.
Specifically, we re-scale the initial weights $W$ of FFNs in the $l$-th layer (larger $l$ means closer to the output) as
\begin{align}
    W^\text{Scaled\_Init} &= \frac{W}{\sqrt{2*(l+1)}}.
\end{align}
Deeper layers are thus initialized with smaller random weights. Recent research has found this type of scaled initialization beneficial for large scale optimization stability~\citep{radford2019language, liu2020understanding}. 

\begin{figure}[t]
\centering
\begin{subfigure}[t]{0.40\textwidth}
\centering
	\includegraphics[width=\textwidth]{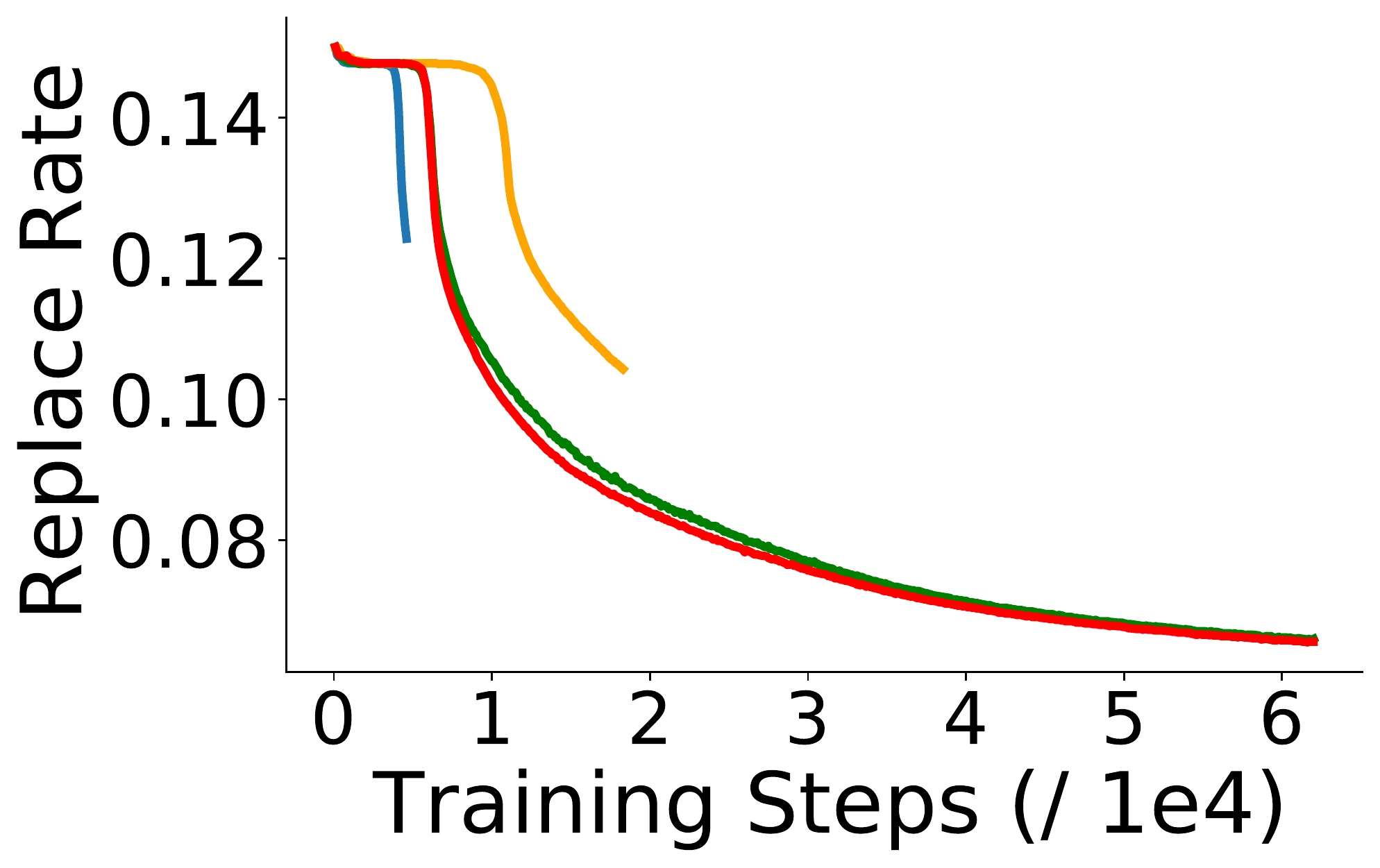}
	\caption{Auxiliary Model\label{fig:dynamicInit_replace_rate}}
\end{subfigure}%
~
\begin{subfigure}[t]{0.40\textwidth}
\centering
	\includegraphics[width=\textwidth]{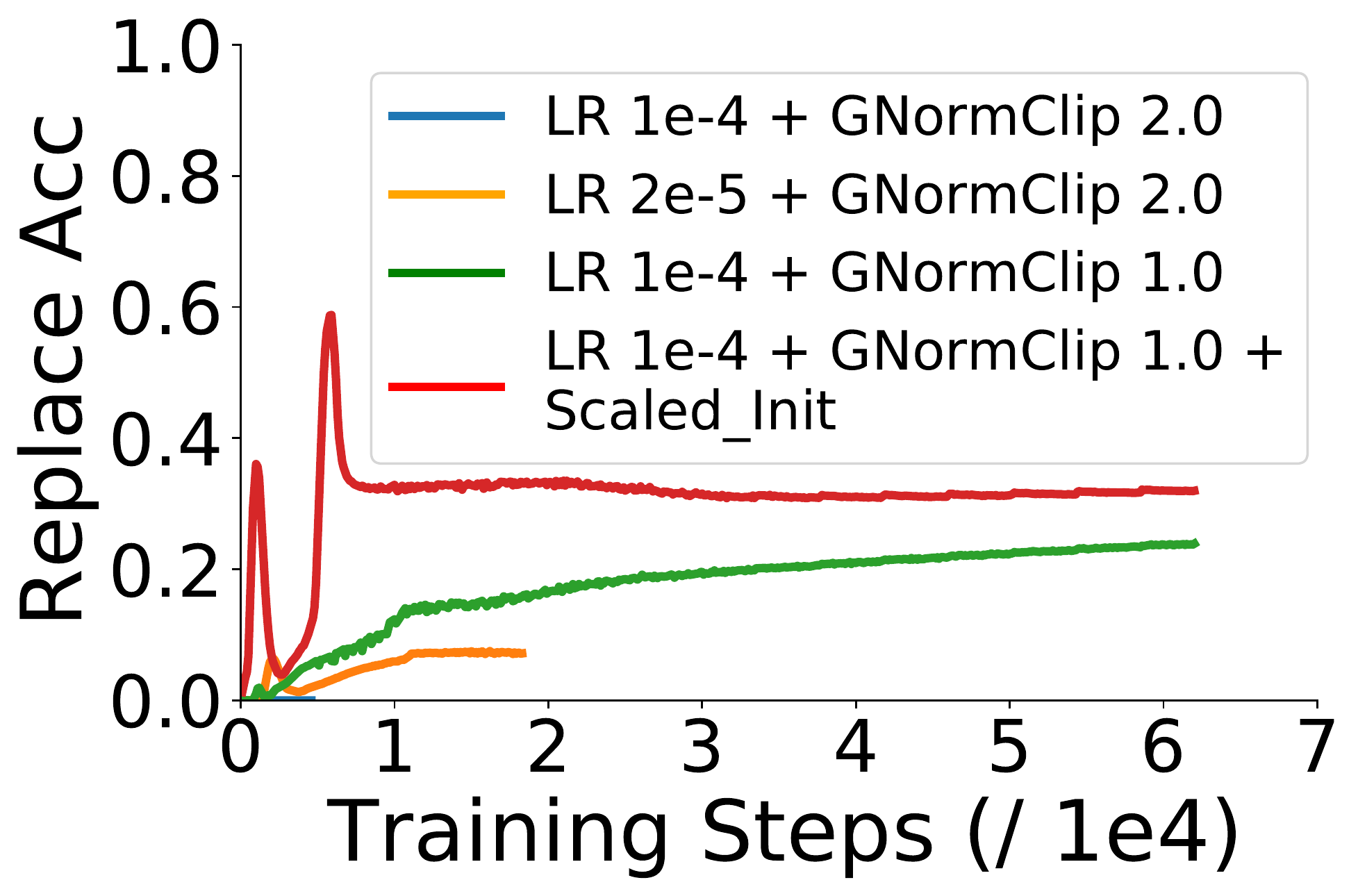}
	\caption{Main Model\label{fig:dynamicInit_replace_acc}}
\end{subfigure}
\caption{Training curves of first 60K pretraining steps of \model{}$_\text{XL}$ with different optimization configurations. Replace Rate is the fraction of tokens replaced by the auxiliary model (out of all tokens).  Replace Accuracy is the fraction of replaced tokens detected by the main model.
}\label{fig:DynamicInit}
\end{figure}

Figure~\ref{fig:DynamicInit} illustrates the pretraining process of \model{}$_\text{XL}$ with different stability techniques.
The Replace Rate is the number of errors the auxiliary model's MLM prediction makes on the 15\% masked out positions. 
All runs start with near 15\% replace rate, a random sampling, and quickly learn to construct more deceiving noises with a lower replace rate.
The Replace Accuracy measures the performance of the RTD task on the replaced tokens, which we found most indicative of the main model's learning process. 
With reduced learning rate and stronger gradient norm (GNorm) clipping, we are able to pretrain \model{}$_\text{XL}$ without divergence. Still, the main Transformer struggles to learn; its replace accuracy is quite low. 
With scaled initialization, the main Transformer, which is much deeper than the auxiliary model (48 versus 8 layers), finally  catches up with the increasingly difficult noise signals generated by the auxiliary model, and yields more stable replace accuracy through the pretraining process.

\subsection{Stable Fine-Tuning}
As language models grow bigger, it is harder to fully finetune them on downstream tasks. The large models have the capability to memorize the training labels.
To improve the generalization and robustness of \model{} on downstream tasks, we utilize Posterior Differential Regularization (PDR)~\citep{cheng2020pdr} for finetuning. This is a simplified version of virtual adversarial training \citep{miyato2018virtual, jiang2019smart, liu2020alum}. PDR promotes local smoothness of the model via stabilizing its posterior distribution towards small input perturbations. 

To finetuning with input embeddings $\bs{x}$ and target $y$, PDR optimizes the model parameter $\Theta$ as:
\begin{align}
\min_\Theta \text{ }  & \mathcal{L}(f_\Theta(\bs{x}), y) + \alpha \mathcal{R}(f_\Theta(\bs{x}), f_\Theta(\hat{\bs{x}})); \\
&\hat{\bs{x}} = \bs{x} + \epsilon; \epsilon\in\RR^n
\end{align}
where $\epsilon$ is a random vector bounded in a norm ball with radius $c$, and $\mathcal{R}$ is a regularization term penalizing the model instability, i.e., the KL divergence between model outputs with and without perturbations. The two terms are balanced by $\alpha$ which we set to $1$.  

In addition, We use Multi-Task Learning (MTL)~\citep{caruana1997multitask,liu2019mt-dnn} by leveraging additional supervised data from other related tasks. Specifically, we use \model as a shared text encoder, stack on top of it a task-specific head for each task, and train these tasks jointly. This facilitates us to experiment data augmentation (DA) on MNLI, training it jointly with SNLI~\citep{snli2015} and ANLI~\citep{nie2020anli} using multi-task learning.

\section{Experimental Methodologies for Large Models}
In this section, we describe the experimental setups,  implementation details, and downstream evaluations for pretraining model at larger scales.

\paragraph{Pretraining Setup.} 
The configurations of \model{} at four different sizes are presented in Table~\ref{tab:model_configs}. 
These \model{} are pretrained using the RoBERTa setting: pretraining for 4B samples (2048 batch size and 2M steps) of 512 length sequences. We use the training corpus 
that consists of 160 GB text from Wikipedia, BookCorpus~\citep{zhu2015aligning}, STORIES~\citep{trinh2018simple}, CC-News~\citep{liu2019roberta}, and OpenWebText~\citep{Gokaslan2019OpenWeb}. 
This pretraining setting is marked by suffix ``++'' when needed, e.g., ``base++'' and ``large++'', to differentiate it from the BERT base setting used in Section~\ref{sec:model}, as the two differentiate significantly on training steps and corpus sizes.

\paragraph{Downstream Evaluation.} We evaluate \model{} under two settings. The first is vanilla finetuning, where we evaluate all \model{} sizes on the dev set of GLUE~\citep{wang2018glue} language understanding benchmark and SQuAD~\citep{rajpurkar2016squad} reading comprehension benchmark. 
This setting finetunes all models individually for each task and reports the median of 5 random seeds. This is a standard evaluation setting used in pretraining research.

The second is to evaluate the best possible performance of \model{} with more advanced finetuning techniques on GLUE and SuperGLUE test sets~\citep{wang2019superglue}, as discussed in past section.
In addition, following \citep{liu2019roberta, liu2019mt-dnn}, we fine-tune other GLUE tasks except CoLA based on the pretrained checkpoint of (XXL-PDR-DA) which is jointly trained by using PDR~\citep{cheng2020pdr} on ANLI~\citep{nie2020anli}, SNLI~\citep{snli2015} and MNLI~\citep{MNLI}. 
We report test results obtained from the GLUE official evaluation server.  More finetuning details can be found in appendix.

\paragraph{Implementation Details.} Our implementations are built upon PyTorch, FairSeq~\citep{ott2019fairseq}, and DeepSpeed \citep{rajbhandari2020zero}, and MT-DNN (for finetuning)~\citep{liu2019mt-dnn, liu2020mtmtdnn}, with A100 GPUs as the main computing hardware. We share more about our pretraining configurations, hyperparameters, and implementation details in appendix.

\begin{table*}[t]
\centering

\small 
\resizebox{\textwidth}{!}{
\begin{tabular}{lll*{9}{l}ll}
\toprule {\textbf{Model}}
 & {\textbf{Total}} & \multicolumn{9}{c}{\textbf{GLUE DEV Single Task}} & \multicolumn{2}{c}{\textbf{SQuAD 2.0}} \\ 
\cmidrule(lr){3-11}\cmidrule(lr){12-13}
 & {\textbf{Params}} & \textbf{MNLI-(m/mm)} & \textbf{QQP} & \textbf{QNLI} & \textbf{SST-2} & \textbf{CoLA} & \textbf{RTE} & \textbf{MRPC} & \textbf{STS-B} & \textbf{AVG} &
\textbf{EM} & \textbf{F1}\\
\midrule
\multicolumn{12}{l}{\textbf{Base++}: RoBERTa base Transformer size and 4B pretraining samples.} \\
\midrule
XLNet~\citep{yang2019xlnet} & 110M
& 86.8/- & 91.4 & 91.7 & 94.7 & 60.2 & 74.0 & 88.2 & 89.5  & 84.6 & 80.2 &--\\
RoBERTa~\citep{liu2019roberta} & 125M
& 87.6/- & 91.9 & 92.8 & 94.8 & 63.6 &  78.7 & 90.2 &  91.2  & 86.4 & 80.5 & 83.7\\
UniLM V2~\citep{unilmv2} & 110M
& 88.5/- & 91.7 & 93.5 & 95.1 & 65.2 & 81.3 & \textbf{91.8}  & 91.0  & 87.1 & 83.3 & 86.1\\
DeBERTa~\citep{he2020deberta} & 134M
&  88.8/88.5 &--&--&--&--&--&--&--&--& 83.1 & 86.2 \\
COCO-LM~\citep{meng2021coco}  & 134M 
& 90.2/90.0
 &	92.2 &	94.2 & 94.6 & 67.3 &	87.4 & 91.2 & 91.8 & 88.6 & 85.4 & 88.1 \\
\model{}$_\text{base}$ & 184M & \textbf{90.3/90.2} & \textbf{92.4}  &   \textbf{94.4}    &	\textbf{95.9}   &	\textbf{71.8}   &	\textbf{88.1}   &	91.4   &	\textbf{92.0}    & \textbf{89.5}   &   \textbf{85.9}	&   \textbf{88.5}\\
\midrule
\multicolumn{12}{l}{\textbf{Large++:}  RoBERTa large Transformer size and 4B pretraining samples.} \\
\midrule 
XLNet~\citep{yang2019xlnet} & 360M
& 90.8/90.8 &	92.3 &	94.9 &	\textbf{97.0} &	69.0 & 85.9 &	90.8 &	92.5	& 89.2 & 87.9 & 90.6 \\
RoBERTa~\citep{liu2019roberta} & 356M
& 90.2/90.2 &	92.2 &	94.7 &	96.4 &	68.0 &	86.6 &	90.9 &	92.4 &	88.9 & 86.5 & 89.4 \\
ELECTRA~\citep{clark2020electra} & 335M
& 90.9/- & 92.4 & 95.0 & 96.9 & 69.1 & 88.0 & 90.8 & 92.6  & 89.4 & 88.0 & 90.6 \\
DeBERTa~\citep{he2020deberta} & 384M
& 91.1/91.1 & 92.3 & 95.3 & 96.8 & 70.5 & -- &-- &--& 88.0 & 90.7 \\ 
COCO-LM~\citep{meng2021coco} & 367M
& 91.4/91.6 & 92.8 & 95.7 & 96.9 & 73.9 & 91.0 & \textbf{92.2} & 92.7  & 90.8 & 88.2 & 91.0 \\ 
\model{}$_\text{large}$  & 434M & \textbf{91.7/91.7}  &	\textbf{92.9}  &	\textbf{95.8} &	96.3 &	\textbf{75.2}    &	\textbf{93.1}    &	\textbf{92.2} &	\textbf{92.8}  & \textbf{91.3} &	   \textbf{88.5} &	\textbf{91.1} \\
\midrule
\multicolumn{12}{l}{\textbf{XL++ and XXL++:} Extra large Transformers.} \\
\midrule
Megatron$_\text{1.3B}$~\citep{shoeybi2019megatron} & 1.3B
 & 90.9/91.0 & 92.6 &--&--&--&--&--&--&--& 87.1 & 90.2 \\
Megatron$_\text{3.9B}$~\citep{shoeybi2019megatron} & 3.9B
& 91.4/91.4 & 92.7 &--&--&--&--&--&--&--& 88.5 & 91.2
\\
DeBERTa~\citep{he2020deberta} & 1.5B & 91.7/91.9 & 92.7 & 96.0 & 97.2 & 72.0 & -- & -- & -- & -- & \textbf{89.7} & 92.2 \\
\model{}$_\text{XL}$ & 1.6B &  92.2/92.0 &	\textbf{93.2} &	96.3    &	\textbf{97.3} &	\textbf{76.0}     &	\textbf{93.5}     &	\textbf{91.7} &	\textbf{93.0}   	& \textbf{91.6}   &		89.4    &	92.1\\ 
\model{}$_\text{XXL}$  & 5.4B & \textbf{92.3/92.4} &	93.1  &  \textbf{96.5}   &	\textbf{97.3}    &	-    &	-    &	- & - & - & \textbf{89.7}	& \textbf{92.4}\\
\midrule
\end{tabular}
}
\caption{
Results on GLUE and SQuAD 2.0 development sets. All results are single-task, single-model fine-tuning. We use Spearman correlation for STS, Matthews correlation for CoLA, and accuracy for the rest on GLUE. Models are grouped into \textit{Base++} and \textit{Large++} based on their Transformer encoder size. The parameters differences are mainly from their vocabulary sizes.
}
\vspace{-0.5em}
\label{tab:model_scale_res}
\end{table*}

\section{Evaluation Results.}
In this section, we first evaluate models effectiveness and their scaling efficiency, then we discuss the advantage of \model{} and room for improvements.

\subsection{Results with Vanilla Fine-tuning}
Table~\ref{tab:model_scale_res} presents the evaluation results with vanilla finetuning. 
At all model sizes, \model{} outperforms all baselines on nearly all tasks. It establishes the new state-of-the-art for GLUE and SQuAD 2.0. More importantly,  \model{} often outperforms previous LLMs at bigger sizes. 
Using only a fraction of parameters, \model{}$_\text{large}$ performs similarly or better than many previous XL and XXL models. 
\model{}$_\text{XL}$ and \model{}$_\text{XXL}$ also achieve the best performances on nearly all tasks evaluated, showing the benefits of our method in terms of both model efficiency and model effectiveness.

Among the ELECTRA-style models, \model{} provides the best combination of effectiveness and efficiency. While using the same encoder configurations in {base} and {large}, \model{} outperforms the original ELECTRA significantly. It also outperforms COCO-LM  most tasks while being cheaper to train. 
METRO is the first recipe that effectively scales up the ELECTRA-style approach to pretraining models with billions of parameters. Our experimental results for the first time demonstrate the advantages of using model generated denoising objectives for pretraining LLMs at the scale of billions parameters.

Table~\ref{tab:model_scale_res} does not report the results on smaller GLUE tasks from several large models, including \model{}$_\text{XXL}$. As models get bigger and bigger, they start to overfit the limited training labels of these tasks. This leads to highly unreliable results. Therefore, more advanced finetuning techniques are required to effectively leverage LLMs when training labels are limited in quantity, as we will discuss next.



\subsection{Results with Stable Fine-Tuning}
\label{subsec:sft}

In this set of experiments, we first study the benefit of stable fine-tuning techniques on the MNLI task, the ``pre-finetuning'' task whose checkpoints are often used as starting points of other language applications, and then show the results of \model{} with stable finetuning.


\begin{table*}[t!]
\center

\resizebox{\textwidth}{!}{
		\begin{tabular}{lll*{9}{l}ll}
        \toprule
         \textbf{Model} & {\textbf{Params}} & \textbf{MNLI-(m/mm)} & \textbf{QQP} & \textbf{QNLI} & \textbf{SST-2} & \textbf{CoLA} & \textbf{RTE} & \textbf{MRPC} & \textbf{STS-B} & \textbf{WNLI} & \textbf{AX}     &\textbf{Score}\\
        \midrule
            \model{}$_\text{XXL}$ (Dev Best)    &         5.4B     &	92.7/93.0 & 91.2/93.4 & 97.5    &97.3 &78.1   &95.7  & 93.9/92.4  & 93.3/93.2        & 97.1	&  - & -  \\ \midrule
            Human Performance      & --    &92.0/92.8 &59.5/80.4 &91.2  &97.8  &66.4    &93.6  &86.3/80.8  &92.7/92.6    	        & \textbf{95.9}	&-      &87.1 \\
            RoBERTa           &  356M   &90.8/90.2  &74.3/90.2  &98.9   &96.7 &67.8  &88.2   &92.3/89.8  &92.2/91.9    	        &89.0	&48.7   &88.5 \\
            MT-DNN-{SMART}      &      350M &91.0/90.8      &73.9/90.2  & \textbf{99.2}   &97.5  &69.5    &89.7 &93.7/91.6  &92.9/92.5   	      & 94.5	& 50.2  & {89.9} \\
            T5$_\text{11B}$ & 11B & 92.2/91.9 & 75.1/90.6 & 96.9 & 97.5 & 71.6 & 92.8 & 92.8/90.4 & 93.1/92.8 & 94.5 & 53.1 & 90.3\\
            ERNIE 3.0 & 10B & 92.3/91.7 & 75.2/90.9 & 97.3 & \textbf{97.8} & \textbf{75.5} & 92.6 & \textbf{93.9/91.8} & 93.0/92.6 & \textbf{95.9} & 51.7 & 91.1\\
            \model{}$_\text{XXL}$    &   5.4B            & \textbf{92.6/92.4} & \textbf{76.4/91.1}  & 97.9    & 97.6 &  72.6   & \textbf{94.1}  & 93.8/91.7  &\textbf{93.7/93.3}   	       & \textbf{95.9}	& \textbf{57.0}  & \textbf{91.2}  \\
            \bottomrule
		\end{tabular}
	}
	\caption{Best dev results from \model{}$_\text{XXL}$ and its GLUE test set results, obtained from the GLUE evaluation server. Advanced fine-tuning techniques are used.\label{tab:glue}
} 
\end{table*}

\begin{table*}[th!]
    \centering

    \small
    \resizebox{\textwidth}{!}{
    \begin{tabular}{lll*{6}{l}ll}
        \toprule
        {\bf Model} & \textbf{Params} & \textbf{BoolQ} &\textbf{CB}&\textbf{COPA}&\textbf{MultiRC}&\textbf{ReCoRD}&\textbf{RTE}&\textbf{WiC}&\textbf{WSC}& \textbf{Average}\\ 
        \midrule
        \model{}$_\text{XXL}$ (Dev Best) & 5.4B & 91.9& 95.7/98.2 & 97.0 &88.0/62.6 &96.2/95.7 &95.7 & 77.0 &  97.1& - \\ \hline

        RoBERTa\textsubscript{large} & 356M &  87.1  & 90.5/95.2 & 90.6 &84.4/52.5&90.6/90.0&88.2&69.9 &89.0 &84.6\\ 
        T5$_\text{11B}$ & 11B & 91.2  & 93.9/96.8 & 94.8 &88.1/63.3&94.1/93.4&92.5&76.9 &93.8 &89.3\\ 
        Human  & -- & 89.0 & 95.8/98.9 & 100.0 &81.8/51.9&91.7/91.3&93.6&80.0 &100.0 &89.8\\ 
        DeBERTa & 1.5B & 90.4  & 95.7/97.6 & 98.4&88.2/63.7&94.5/94.1&93.2&77.5 &	95.9 & 90.3\\ 
        ERNIE 3.0& 10B &  91.0 & \textbf{98.6/99.2} & 97.4 & 88.6/63.2 &94.7/94.2 &92.6 & 77.4 & 97.3 & 90.6 \\
        ST-MoE-32B& ``32B'' &  \textbf{92.4} & 96.9/98.0 & \textbf{99.2} & \textbf{89.6/65.8} &95.1/94.4 &93.5 & \textbf{77.7} & 96.6 & \textbf{91.2} \\
        \model{}$_\text{XXL}$ & 5.4B &  92.0 & 95.9/97.6 & 98.2 & 88.4/63.0 &\textbf{96.4/95.9} &\textbf{94.1} & 77.1 & \textbf{97.3} & 90.9 \\
            \bottomrule
		\end{tabular}
	}
	    \caption{Best dev results from \model{}$_\text{XXL}$ and its SuperGLUE test set results, obtained from the SuperGLUE evaluation server. Advanced fine-tuning techniques are used. The parameters of MoE models are the potential size of dense models with equivalent inference cost. 
	    \label{tab:superglue} 
    } 
\end{table*}

\begin{figure}[t]
    \centering
    \includegraphics[width=0.5\linewidth]{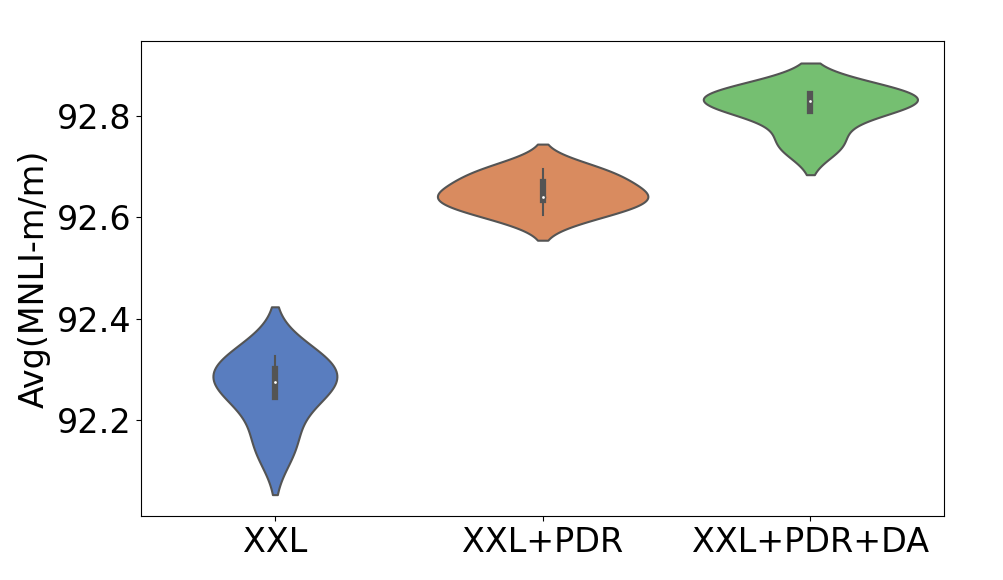}
    
    \caption{Score distribution on MNLI-m/mm dev average of \model{}$_\text{XXL}$ with and without Posterior Differential Regularization (PDR) and Data Augmentation (DA).   \label{fig:ft_stragety} }
\end{figure}

Figure~\ref{fig:ft_stragety} plots the score distribution of \model{}$_\text{XXL}$ with different techniques. Each result is averaged over 5 random runs in the same setting. PDR improves both average accuracy and stability, and yields 0.4 accuracy boost on MNLI. Data augmentation by adding ANLI and SNLI tasks to MNLI in multi-task learning further boosts the performance. Combining the two techniques significantly improves the effectiveness on MNLI, providing a strong ``pre-finetuned' starting point for other GLUE and SuperGLUE tasks.

We then apply the single-task PDR learning technique to continuously finetune on other tasks, obtain the best \model{}$_\text{XXL}$ models on Dev, and submit them to GLUE and SuperGLUE test servers to perform blind evaluation. The results with stable finetuning are listed in Table~\ref{tab:glue} and Table~\ref{tab:superglue}. The empirical advantage of \model{} on Dev holds on Test. \model{}$_\text{XXL}$ achieved new state-of-the-art results on both leaderboards. It is the first model that achieves accuracy that surpasses the estimated human performance on MNLI and RTE, the only two remaining GLUE tasks where human level performance had not been achieved before.

The comparison of \model{}$_\text{XXL}$ with other LLMs on the test sets demonstrates the scaling efficiency of our technique. With 5.4 billion parameters, \model{}$_\text{XXL}$ outperforms many previous models consisting of 10 billions parameters or more, such as ERNIE 3.0 and T5$_\text{11B}$. Our model also outperforms ST-MoE-32B, the recent state-of-the-art mixture-of-experts (MoE) model on several SuperGLUE tasks. The latter uses 64 experts, each equal to a dense 32 billion parameter model, nearly six times bigger than \model{}$_\text{XXL}$. The experiment in the next section further studies the scaling efficiency of \model{}.

\begin{figure*}[t]
\centering
\begin{subfigure}[t]{0.24\textwidth}
\centering
	\includegraphics[width=\textwidth]{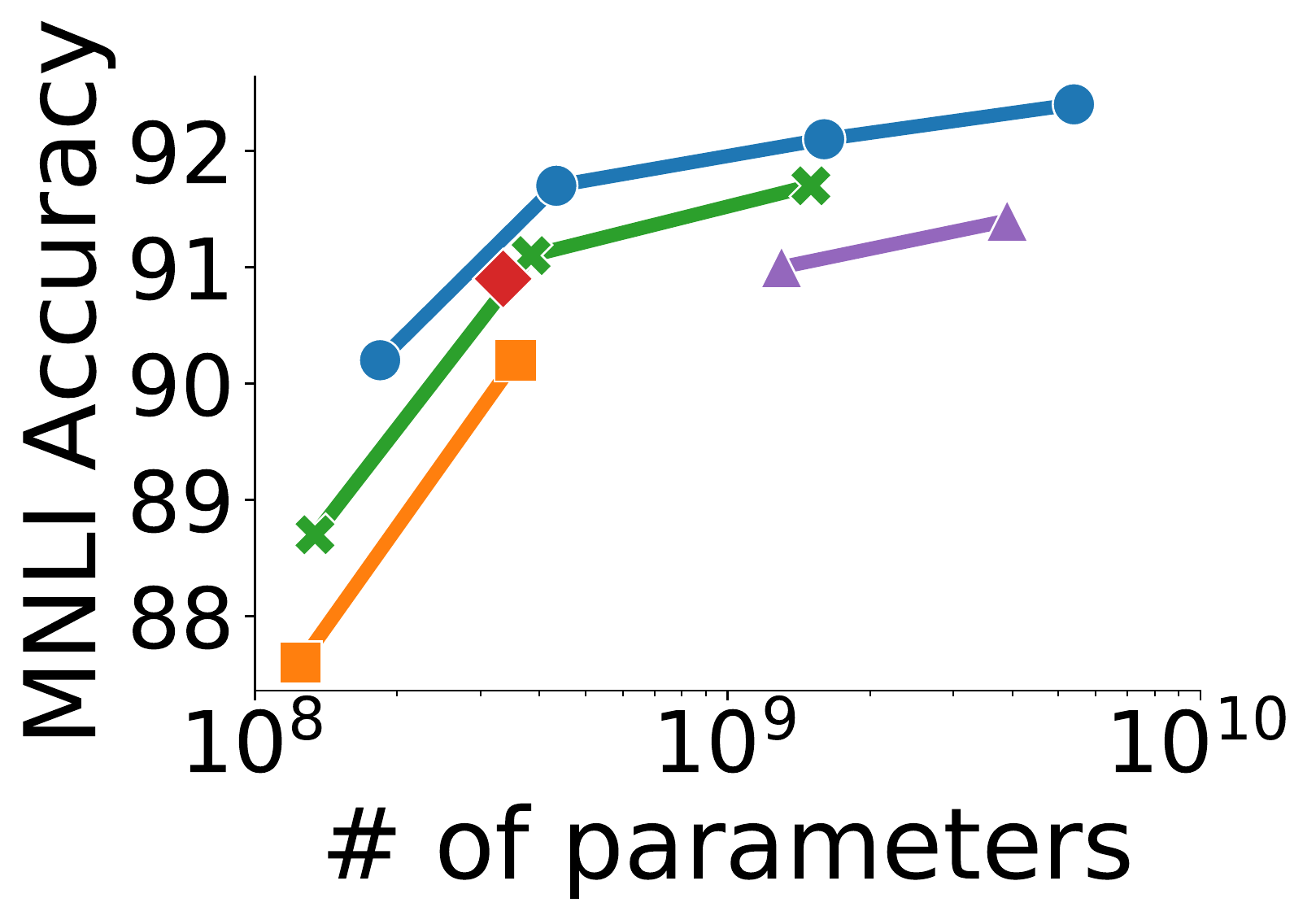}
	\caption{MNLI w.r.t. Size. \label{fig:mnli_model}}
\end{subfigure}%
~
\begin{subfigure}[t]{0.24\textwidth}
\centering
	\includegraphics[width=\textwidth]{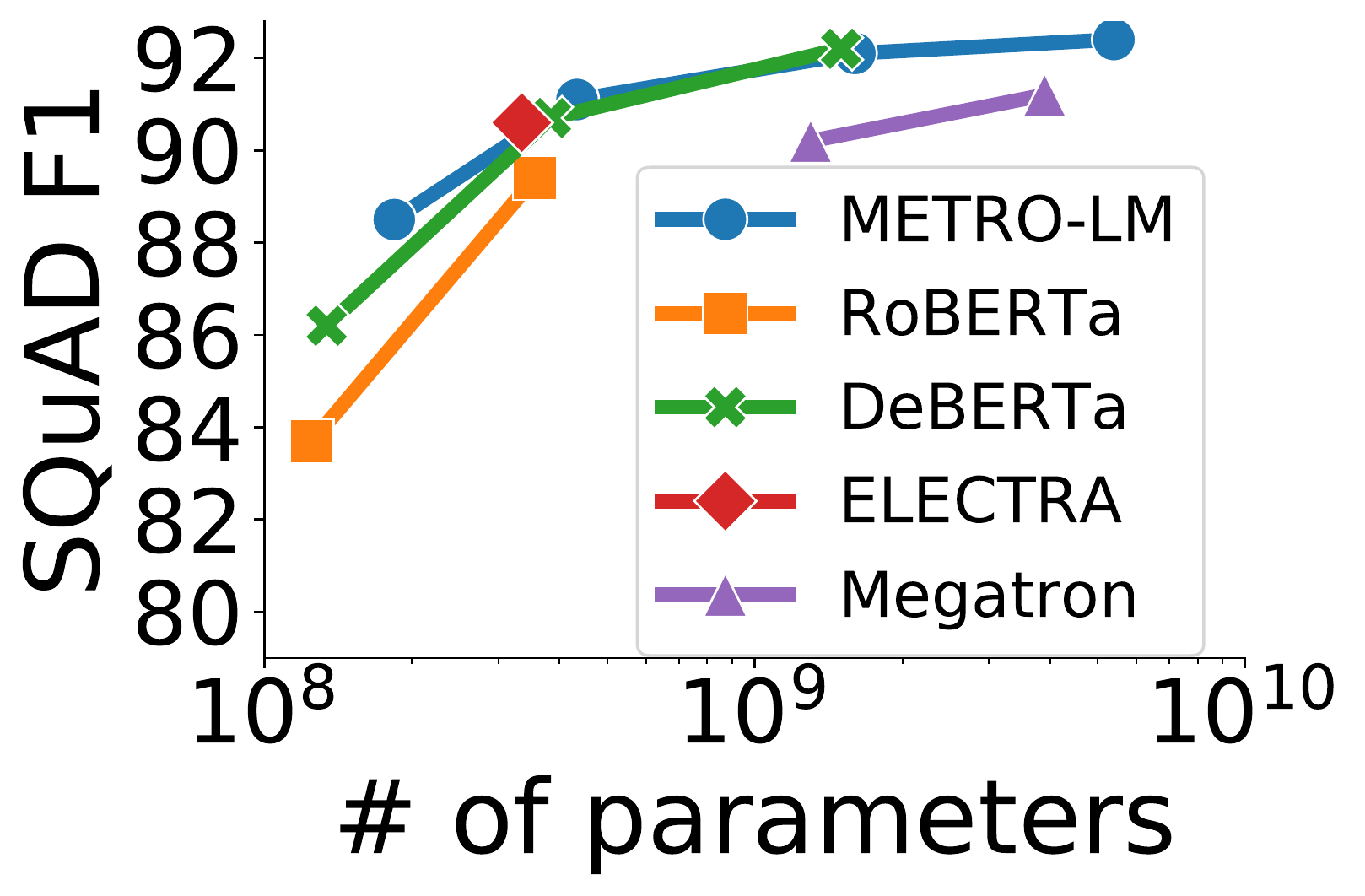}
	\caption{SQuAD w.r.t. Size.  \label{fig:squad_model}}
\end{subfigure}%
~
\begin{subfigure}[t]{0.24\textwidth}
\centering
	\includegraphics[width=\textwidth]{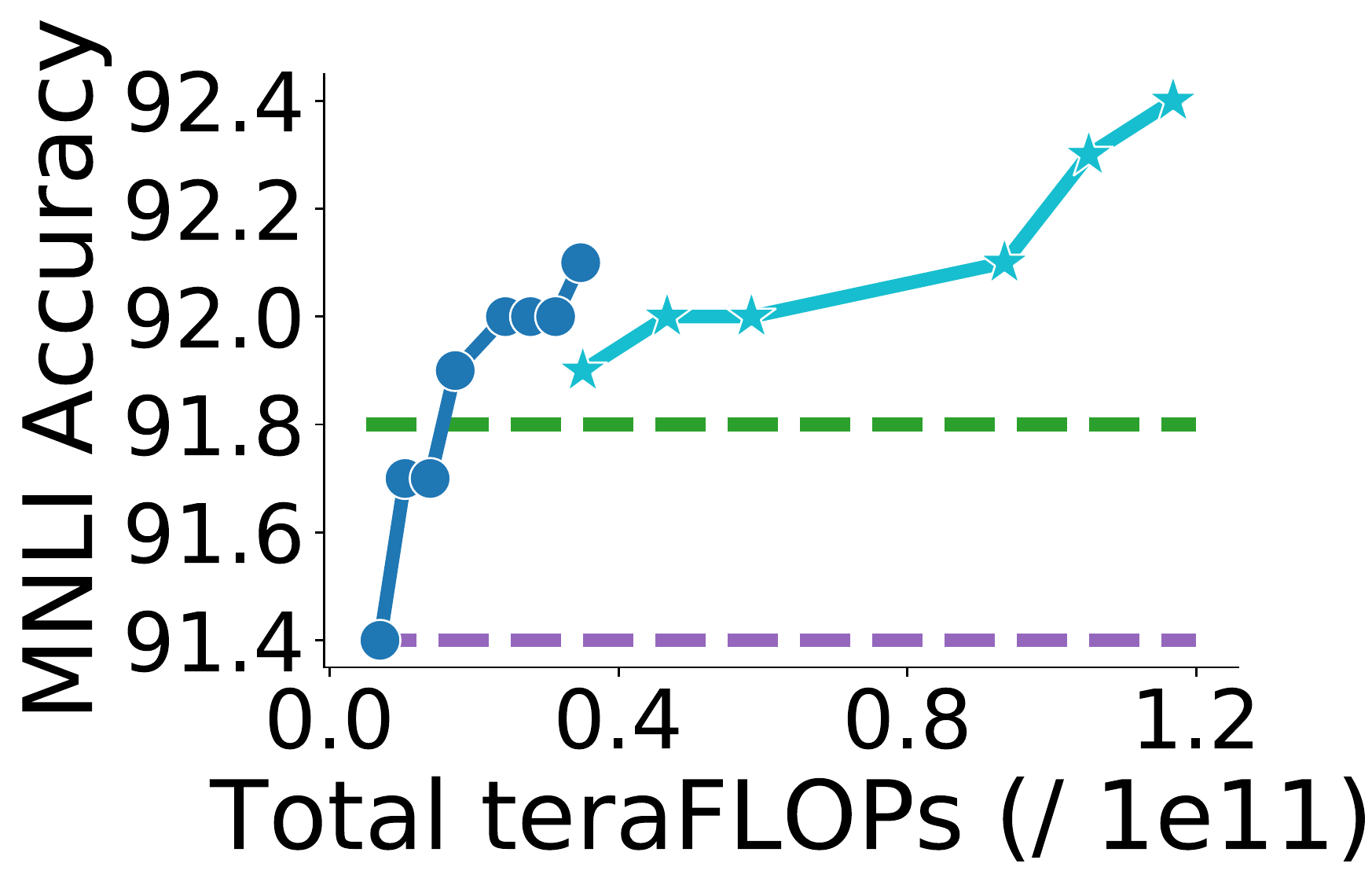}
	\caption{MNLI w.r.t. FLOPs.\label{fig:mnli_time}}
\end{subfigure}%
~
\begin{subfigure}[t]{0.24\textwidth}
\centering
	\includegraphics[width=\textwidth]{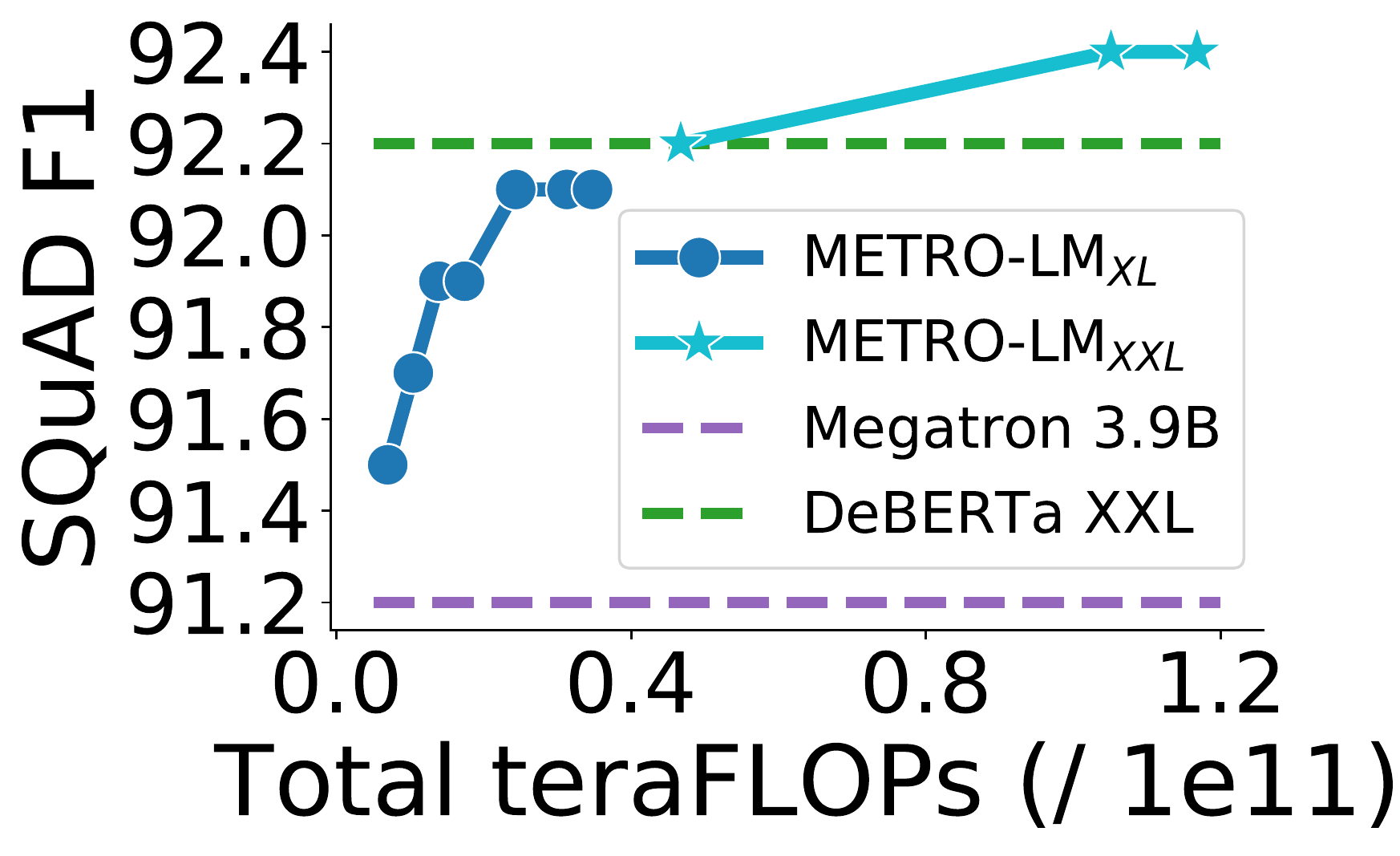}
	\caption{SQuAD w.r.t. FLOPs. \label{fig:squad_time}}
\end{subfigure}
\caption{Model performances on the dev set of MNLI-m and SQuAD F1 w.r.t. model sizes, marked in logarithmic by x-axes (a, b), and at different pretraining computing cost (total teraFLOPs at different percentages of pretraining steps) on x-axes of (c, d).  
\vspace{-1em}
}\label{fig:mnli_squad_results}
\end{figure*}
\subsection{Scaling Efficiency}
Figure~\ref{fig:mnli_squad_results} shows the scaling efficiency of \model{} regarding model size and computing cost.
The ``smaller'' \model{} variants show strong result on MLNI and SQuAD. For example, \model{}$_\text{base}$ achieves a comparable performance to RoBERTa$_\text{large}$, and \model{}$_\text{large}$ outperforms many recent LLMs containing billions of parameters, both echoing the effectiveness of model generated denoising objectives for LLMs pretraining.
At the sizes of {XL} and {XXL}, we also observe consistent improvement compared to previous billion-scale autoencoders.

In Figure~\ref{fig:mnli_time} and~\ref{fig:squad_time}, we present the finetuning results of the intermediate model checkpoints of \model{}$_\text{XL}$ and \model{}$_\text{XXL}$, respectively, at different pretraining steps. Following \citet{smith2022MTNLG}, we calculate the teraFLOPs used during the pretraining process and mark the computing operations used at these intermediate (and final ones in the rightmost of each curve) checkpoints on the x-axes.  Also, to reduce fine-tuning cost which is significant higher as models grow bigger, we restricted the hyper-parameter search space and only checked several intermediate checkpoints. Though previous large scale autoencoder models in general do not report total teraFLOPs, we report total teraFLOPs for future references.

The computing efficiency of ELECTRA-style models observed at the base model size~\citep[e.g.,][]{clark2020electra, meng2021coco} is also observed at {XL} and {XXL}. \model{}$_\text{XXL}$ outperforms most previous LLMs with only 40\% of pretraining steps.
In addition, we also observe significant improvements after 50\% of pretraining steps, showing the effect of curriculum learning using increasingly difficult pretraining signals.

\subsection{Discussions and Future Work} 

The efficiency and effectiveness of \model{} come from the combination of modeling techniques and scalable training techniques.  On the modeling side, we use the denoising training objectives constructed by auxiliary models~\citep{clark2020electra}, while also explore and integrate recently proposed training objectives~\citep{meng2021coco} and neural architectures~\citep{ke2020tupe}.
On scalable training, we utilized ZeRO optimizer~\citep{rajbhandari2020zero}, fused operations, and a simple scaled initialization method, all of which jointly improved the model training efficiency and stability~\citep{liu2020understanding}.
However, despite the significant advantages of \model{}, there are still a lot of directions for improvement, among which we found the following two most unique and challenging in pretraining LLMs.

\paragraph{Hyperparameter Tuning for Billion-parameter Models.} 
Since it is impossible to search optimal hyperparameters for LLMs via trial and error, we often have to pick a set of hyperparamters and pretrain a large model with dedicated computing resource, hoping that we are lucky enough that the picked one is reasonably good.
An efficient way to tune hyperparameters via learning is highly desirable.
For example, recently, ~\citet{yang2021tuning} proposed a method of mapping the hyperparameters optimized on a small proxy model to that of a much larger target model, which is a promising direction to make hyperparameter selection more effective for billion-parameter models.

\paragraph{Training Stability.} 
This is one of the major bottlenecks for large scale training. In our experiments, models of the XL and XXL sizes show different behaviors due to training instability in comparison with the models that contains ``only'' a few hundred millions of parameters. Moreover, the fact that much instability of training XL and XXL models only happens at the late stage of pretraining makes the problem extremely hard---and costly---to resolve. 
Thus, it is crucial to develop new techniques that allow us to monitor the model training process and adjust the optimization accordingly without significantly increasing training overhead~\citep{liang2022no}.

\section{Conclusions}

We propose a new recipe, METRO, for efficiently pretraining of large scale autoencoding language models. METRO significantly extends the ELECTRA-style denoising pretraining by incorporating a suite of techniques, ranging from Transformer architectures, training objectives, to efficient, stable, and scalable large scale optimization methods.   

We present in detail our construction of the METRO recipe using a comprehensive empirical study of various recent techniques. 
We start with a base model trained using the ELECTRA method and improve it by revising model architecture and exploring new training objectives. Finally, we describe how we scale the models to billions of parameters by integrating various optimization techniques, which significantly improve the efficiency and stability of large scale training. 

Using METRO, we have pretrained a family of autoencoding models, \model{}, the largest of which consists of 5.4 billion parameters. \model{} achieves new state-of-the-art on the GLUE, SuperGLUE, and SQuAD benchmarks. More importantly, \model{}s are efficient in that they often outperform previous LLMs with significantly smaller model sizes and lower pretraining cost.

We believe our observations and insights will facilitate future research in language model pretraining, especially as the advancements of computing hardware, infrastructure, and techniques are making pretraining research feasible with commodity computing infrastructures.

\section*{Acknowledgments}
We would like to acknowledge Li Dong for the scaled initialization method and  Jeff Rasley for DeepSpeed support.

\vskip 0.2in
\bibliography{tacl2021}

\newpage

\clearpage
\appendix
\appendix

\section{Task Formulation in Fine-Tuning}
\label{subsec:task}
In this section we share more details on the fine-tuning experiments.

\textbf{GLUE.} We use the standard task formulation on GLUE as suggested in \citet{wang2018glue}: MNLI, SST-2, MRPC, QQP, QNLI, RTE and WNLI as classification; STS-B as regression. 

Table~\ref{tab:glue_stat} summarizes the GLUE tasks. More details are as follows.

\texttt{MNLI:} Multi-genre Natural Language Inference~\citep{MNLI} contains $393$K training examples obtained via crowdsourcing. The task is to predict whether a given premise sentence entails, contradicts, or is neutral to a given hypothesis sentence. 

\texttt{QQP:} Question Pairs~\citep{QQP} contains $364$K training examples from the Quora question-answering website. The task is to determine whether a pair of questions asked are semantically equivalent.

\texttt{QNLI:} Question Natural Language Inference contains $108$K training examples derived from the Stanford Question Answering Dataset (SQuAD)~\citep{rajpurkar2016squad}. The task is to predict whether a given sentence contains the answer to a given question sentence.

\texttt{SST-2:} Stanford Sentiment Treebank~\citep{SST-2} contains $67$K training examples extracted from movie reviews with human-annotated sentiment scores. The tasks is to determine if the sentence has positive or negative sentiment. 

\texttt{CoLA:} Corpus of Linguistic Acceptability~\citep{COLA} contains $8.5$K training examples from books and journal articles on linguistic theory. The task is to determine whether a given sentence is linguistically acceptable or not. 

\texttt{RTE:} Recognizing Textual Entailment~\citep{RTE-1,RTE-2,RTE-3,RTE-5} contains $2.5$K training examples from textual entailment challenges. The task is to predict whether a given premise sentence entails a given hypothesis sentence or not.

\texttt{MRPC:} Microsoft Research Paraphrase Corpus~\citep{MRPC} contains $3.7$K training examples from online news sources. The task is to predict whether two sentences are semantically equivalent or not.

\texttt{STS-B:} Semantic Textual Similarity~\citep{STS-B} contains $5.8$K training examples
drawn from multiple sources with human annotations on sentence pair semantic similarity. The task is to predict how semantically similar two sentences are on a $1$ to $5$ scoring scale.
\begin{table}[t]
\small
\centering
\begin{tabular}{llllll}
\toprule
 & \textbf{Size} & \textbf{Task} & \textbf{Metric(s)} & \textbf{Domain} \\ 
\midrule
MNLI & 393K & Inference & Accuracy & Misc. \\
QQP & 364K & Similarity  & Accuracy/F1 & Social QA \\
QNLI & 105K & QA/Inference  & Accuracy & Wikipedia \\
SST-2 & 67K & Sentiment  & Accuracy & Movie Reviews \\
CoLA & 8.5K & Acceptability & Matthews Corr. & Misc.\\
RTE & 2.5K & Inference  & Accuracy & Misc.\\
MRPC & 3.7K & Paraphrase & Accuracy/F1 & News\\
STS-B & 7K & Similarity & Pearson/Spearman. & Misc.\\
\bottomrule
\end{tabular}
\caption{Summarizes of tasks included in GLUE benchmark.}
\label{tab:glue_stat}
\end{table}

MNLI is directly fine-tuned from pre-trained {\model} via PDR \citep{cheng2020pdr} and multi-task data augmentation~\citep{liu2019mt-dnn}, often denoted as \textbf{MNLI+}. SST-2, MRPC, QQP, QNLI and RTE tasks are fine-tuned with PDR starting from the converged \textbf{MNLI+}  model. 

\begin{table}[t]
\small
\centering
\begin{tabular}{llllll}
\toprule
 & \textbf{Size} & \textbf{Task} & \textbf{Metric(s)} & \textbf{Domain} \\ 
\midrule
BoolQ & 9.4K & QA & Accuracy & Google queries, Wikipedia \\
CB & 250 & Inference  & Accuracy/F1 & Misc. \\
MultiRC & 5.1K & QA  & F1$_a$/EM & Misc. \\
COPA & 400 & QA  & Accuracy &  Blogs, photography encyclopedia \\
ReCORD & 101K & QA & F1/EM & News (CNN, Daily Mail) \\
WiC & 6K & Word Sense Disambiguation  & Accuracy &  WordNet, VerbNet, Wiktionary\\
RTE & 2.5K & Inference & Accuracy & News, Wikipedia\\
WSC & 554 & Coreference resolution & Accuracy &  fiction books.\\
\bottomrule
\end{tabular}
\caption{The list of tasks included in SuperGLUE benchmark.}
\label{tab:superglue_stat}
\end{table}
\begin{table*}[t]
\small
    \centering
    \resizebox{\textwidth}{!}{
    \begin{tabular}{ l   c   c    c    c  }
        \toprule
         \textbf{Hyper-parameters} & \textbf{Base\&Base++} & \textbf{Large++} &  \textbf{XL} & \textbf{XXL} \\
        \hline
        Dropout of task layer & 0.1 & 0.1 &  \{0.1, 0.15\}& \{0.1,0.15\}  \\
        Warmup Steps & \{50,100,500\} & \{50,100,500\} &   \{50,100,500\}& \{50,100,500\} \\
        Learning Rates &  \{1e-5, 3e-5, 5e-5\}
 & \{1e-6, 5e-6, 1e-5\} &  \{8e-7, 8e-6, 1e-5\}& \{8e-7, 8e-6, 1e-5\}  \\
        Batch Size & \{16,32\} & \{16,32\} &   \{16,32,64\}& \{32, 64, 128\}  \\
        Weight Decay & 0.01 & 0.01 &   0.01  & 0.01\\
        Maximum Training Epochs & 10 & 10 &   10& 10 \\
        Learning Rate Decay & Linear & Linear &   Linear & Linear  \\
        Gradient Clipping & 1.0 & 1.0 &   1.0 & 1.0 \\
        \bottomrule
        \end{tabular}
        }
    \caption{
    Hyper-parameters for fine-tuning \model on GLUE, SuperGLUE and SQuAD 2.0. 
    }
     \label{tbl:ft-hyper}
\end{table*}

\textbf{SuperGLUE.} We also mainly follow the suggested setup from \citep{wang2019superglue} with a few modifications: 

\texttt{BoolQ}, \texttt{CB} and \texttt{MultiRC}: We use the same input format as \citet{wang2019superglue, liu2019roberta} and fine-tune from the \textbf{MNLI+} model. 

\texttt{COPA}: Following \citet{liu2019roberta}, we concatenate the premise and each alternative with \textit{because} and \textit{so} markers instead of \textit{cause} and \textit{effect} question, respectively. 

\texttt{ReCoRD}: We formulate it as a pairwise ranking task with one negative and positive entity for each (passage, query) pair, following \citet{liu2019roberta}. In evaluation, we pick the entity with the highest score for each question. Empirically, it obtains around 2 points of the F1 score improvement over the SQuAD-style span extraction~\citep{rajpurkar2016squad}. 
     
\texttt{WiC}: We concatenate the marked word and the pair of sentence as input and only use the representation of [CLS] token for classification.
     
\texttt{RTE and WSC}: These two tasks are shared with GLUE and we use the same model on them.

Compared to GLUE where all tasks are formulated in the sequence learning format, SuperGLUE tasks have more variations in their formulation---and we found different task formulations, even among the standard ones, do have a decent impact on the final results on SuperGLUE tasks. To facilitate more standardized comparison, we have released our code and pipeline for SuperGLUE fine-tuning.\footnote{https://github.com/namisan/mt-dnn/tree/master/experiments/superglue.} We also list the detailed fine-tuning hyperparameters in Table~\ref{tbl:ft-hyper}.

\begin{table*}[t]
\small
\centering
\resizebox{\textwidth}{!}{
\begin{tabular}{l*{9}{c}}
\toprule
\textbf{Model} & \textbf{Wiki+Book} & \textbf{OpenWebText} & \textbf{Stories} & \textbf{CC-News} & \textbf{Gigaword} &  \textbf{Common Crawl} & \textbf{ClueWeb} & \textbf{C4} & \textbf{Reddit + Discovery} \\
\midrule
BERT & \checkmark  \\
RoBERTa & \checkmark & \checkmark & \checkmark & \checkmark \\
ELECTRA & \checkmark &  &  & & \checkmark & \checkmark & \checkmark \\
DeBERTa & \checkmark & \checkmark & \checkmark & \checkmark \\
Megatron & \checkmark & \checkmark & \checkmark & \checkmark  \\
T5 &  & &  & & & & & \checkmark \\
ERNIE & \checkmark &  & &  & & & & & \checkmark \\
\model & \checkmark & \checkmark & \checkmark & \checkmark  \\
\bottomrule
\end{tabular}
}
\caption{Pretraining Data Corpus Comparison \model}
\label{tab:pretrain_data}
\end{table*}

\begin{table*}[t]
\small
\centering
\resizebox{1\textwidth}{!}{
\begin{tabular}{l*{5}{c}}
\toprule
\textbf{Parameters} & \textbf{Base} & \textbf{Base++} & \textbf{Large++} & \textbf{XL} & \textbf{XXL} \\
\midrule
Max Steps & 125K & 1.95M & 1.95M & 1.95M & 1.95M \\
Peak Learning Rate & 5e-4 & 1.75e-4 & 1e-4 & 5e-5 & 3e-5 \\
Batch Size & 2048 & 2048 & 2048 & 2048 & 2048 \\
Warm-Up Steps & 10K & 10K & 10K & 30K & 30K \\
Sequence Length & 512 & 512 & 512 & 512 & 512 \\
Relative Position Encoding Buckets & 64 & 64 & 128 & 512 & 512 \\
Relative Position Encoding Max Distance & 128 & 128 & 256 & 512 & 512 \\
Loss multiplier $\lambda$  & 50 & 50 & 50 & 50 & 50\\
Adam $\epsilon$ & 1e-6 & 1e-6 & 1e-6 & 1e-6 & 1e-6\\
Adam ($\beta_1$, $\beta_2$) & (0.9, 0.98) & (0.9, 0.98) & (0.9, 0.98) & (0.9, 0.98) & (0.9, 0.98)\\
Clip Norm & 2.0 & 2.0 & 2.0 & 1.0 & 1.0 \\
Dropout & 0.1 & 0.1 & 0.1 & 0.1 & 0.1 \\
Weight Decay & 0.01 & 0.01 & 0.01 & 0.01 & 0.01 \\
\bottomrule
\end{tabular}
}
\caption{Hyperparameters used in pretraining \model}
\label{tab:hp_pretrain}
\end{table*}

\section{More Pretraining Details}

The detailed pre-training data corpus used in the ``++'' settings are listed in Table~\ref{tab:pretrain_data}. We use the same pretraining corpora as in COCO-LM~\citep{meng2021coco} in the ``++'' setting. The \textit{base} experiments use Wiki+Book.
The pre-training hyperparameters used are listed in Table~\ref{tab:hp_pretrain}. 
We follow the standard hyperparameter settings for \textit{base++} and \textit{large++}.
For larger sizes, we try to reuse most hyperparameters in \textit{large++} and mainly adapt those that related to optimization stability. Tuning hyperparameters thoroughly at that scale is costly. We  manually searched a handful of settings till reaching a stable training process with a setting and then use it for the full pretraining run.
For example, we reduced the peak learning rate on \textit{XL} and \textit{XXL}, as well as the gradient clip norm, and increased the warm-up steps to avoid divergence. 
These hyperparameters are unlikely optimal and we expect more future improvements on this front.

\end{document}